\documentclass{article}

\usepackage[margin=1in]{geometry}
\usepackage[table]{xcolor}

\usepackage{graphicx}
\usepackage{booktabs}
\usepackage{amsmath}
\usepackage{amssymb}
\usepackage{amsfonts}
\usepackage{multirow}
\usepackage{verbatim}
\usepackage{caption}
\usepackage{longtable}
\usepackage{supertabular}
\usepackage{float}
\usepackage{lscape}
\usepackage{fancyhdr}

\usepackage{tablefootnote}
\usepackage[round,semicolon]{natbib}
\usepackage{xspace}
\usepackage{textcomp}
\usepackage{makecell}
\usepackage{siunitx}
\usepackage{colortbl}

\newcommand{\emoji}[1]{\includegraphics[height=1em]{emojis/#1}}

\usepackage{pifont}
\definecolor{darkerGreen}{RGB}{0,170,0}
\newcommand\greencheck{\textcolor{darkerGreen}{\ding{52}}}

\usepackage[]{mdframed}

\usepackage[subfigure]{tocloft}

\usepackage{scrextend}

\usepackage{array}
\usepackage{tgpagella}
\usepackage{latexsym}
\usepackage[T1]{fontenc}
\usepackage{microtype}
\definecolor{mydarkblue}{rgb}{0,0.08,0.45}
\PassOptionsToPackage{hyphens}{url}\usepackage[colorlinks,citecolor=mydarkblue,urlcolor=mydarkblue,linkcolor=mydarkblue,linktocpage=true]{hyperref}
\usepackage{url}            
\usepackage{nicefrac}       
\usepackage{changepage}
\usepackage{xargs}          
\usepackage{wrapfig,lipsum,booktabs}
\usepackage{subcaption}
\usepackage{endnotes}
\usepackage{adjustbox}

\usepackage{pgfplots}
\usetikzlibrary{pgfplots.groupplots}
\pgfplotsset{compat=1.3}
\usepackage{tikz}
\usetikzlibrary{patterns}

\usepackage[most]{tcolorbox}

\usepackage[capitalize,noabbrev]{cleveref}
\crefname{section}{Section}{\S\S}
\Crefname{section}{Section}{\S\S}
\crefname{table}{Table}{Tables}
\crefname{figure}{Figure}{Figures}
\crefname{algorithm}{Algorithm}{}
\crefname{equation}{eq.}{}
\crefname{appendix}{Appendix}{}
\crefformat{section}{Section #2#1#3}
\usepackage{multicol}

\usepackage{xspace}
\usepackage[textsize=tiny]{todonotes}

\newcommand{\dpiex}[0]{DPExplorer}
\newcommand{\dpic}[0]{DPCollection}
\newcommand{\ncao}[0]{NC/A-O\xspace}

\newlength{\myMheight}
\newcommand{\github}{\includegraphics[height=\myMheight]{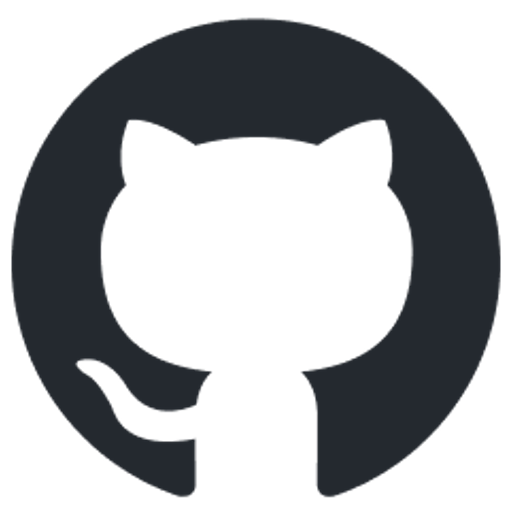}}
\newcommand{\pwc}{\includegraphics[height=\myMheight]{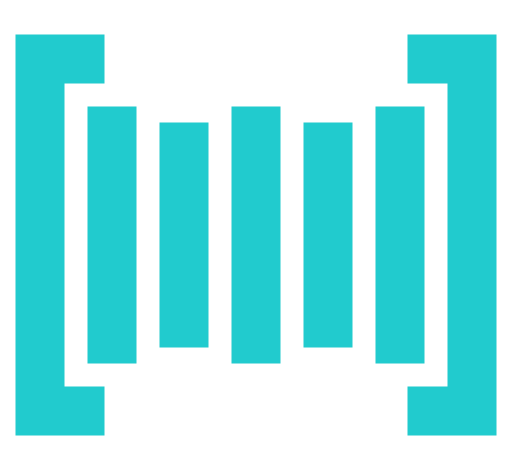}}
\newcommand{\hf}{\includegraphics[height=\myMheight]{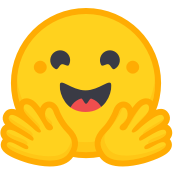}}

\usepackage{minitoc}

\usepackage[bottom,symbol]{footmisc}

\vspace{-10mm}
\title{
    \textbf{
    The Data Provenance Initiative:\\A Large Scale Audit of Dataset Licensing \& Attribution in AI
    }
}
\vspace{-3mm}

\renewcommand\footnotemark{}
\newcommand\tinyspace{\hspace{0.12em}}

\author{
    \normalsize{}
    \textbf{Shayne Longpre\tinyspace$^{\mathbf{1}\tinyspace\text{\textdagger}}$\thanks{\textsuperscript{\textdagger}\,Correspondence: \url{data.provenance.init@gmail.com}}}
    \hspace{4mm}
    \textbf{Robert Mahari\tinyspace$^{\mathbf{1,2}}$}
    \hspace{4mm} 
    \textbf{Anthony Chen\tinyspace$^{\mathbf{3}}$}
    \hspace{4mm}
    \textbf{Naana Obeng-Marnu\tinyspace$^{\mathbf{1,4}}$}
    \\
    \normalsize{}
    \textbf{Damien Sileo\tinyspace$^{\mathbf{5}}$}
    \hspace{4mm} 
    \textbf{William Brannon\tinyspace$^{\mathbf{1,4}}$}
    \hspace{4mm}
    \textbf{Niklas Muennighoff\tinyspace$^{\mathbf{6}}$}
    \hspace{4mm} 
    \textbf{Nathan Khazam\tinyspace$^{\mathbf{7}}$}
    \\
    \normalsize{}
    \textbf{Jad Kabbara\tinyspace$^{\mathbf{1,4}}$}
      \hspace{4mm}
    \textbf{Kartik Perisetla}
    \hspace{4mm}
    \textbf{Xinyi (Alexis) Wu\tinyspace$^{\mathbf{8}}$}
    \hspace{4mm}
    \textbf{Enrico Shippole}
    \hspace{4mm} 
    \textbf{Kurt Bollacker\tinyspace$^{\mathbf{7}}$}
    \\
    \normalsize{}
    \textbf{Tongshuang Wu\tinyspace$^{\mathbf{9}}$}
    \hspace{4mm}
    \textbf{Luis Villa\tinyspace$^{\mathbf{10}}$}
    \hspace{4mm}
    \textbf{Sandy Pentland\tinyspace$^{\mathbf{1}}$}
    \hspace{4mm}
    \textbf{Sara Hooker\tinyspace$^{\mathbf{11}}$}
    \\
    \\
    \normalsize{}
    $^{1}$ MIT
    \hspace{4mm}
    $^{2}$ Harvard Law School
    \hspace{4mm}
    $^{3}$ UC Irvine
    \hspace{4mm}    
    $^{4}$ MIT Center for Constructive Communication
    \\
    \normalsize{}
    $^{5}$ Inria, Univ. Lille Center
    \hspace{4mm}
    $^{6}$ Contextual AI
    \hspace{4mm}
    $^{7}$ ML Commons
    \hspace{4mm}
    $^{8}$ Olin College
    \\
    \normalsize{}
    $^{9}$ Carnegie Mellon University
    \hspace{4mm}
    $^{10}$ Tidelift
    \hspace{4mm}
    $^{11}$ Cohere For AI
    \vspace{-10mm}
}

\definecolor{ncred}{HTML}{e04c71}
\definecolor{unspecgold}{HTML}{e0cd92}
\definecolor{cblue}{HTML}{82b5cf}

\newcommand{\emojiblank}{\phantom{\emoji{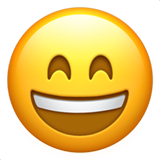}}}
\newcommand{\NCDataCircle}{\tikz[baseline=-0.85ex]{\fill[ncred] (0,0) circle (0.85ex);}}
\newcommand{\UnspecifiedDataCircle}{\tikz[baseline=-0.85ex]{\fill[unspecgold] (0,0) circle (0.85ex);}}
\newcommand{\CommercialDataCircle}{\tikz[baseline=-0.85ex]{\fill[cblue] (0,0) circle (0.85ex);}}
\newcommand{\TransparentCircle}{\tikz[baseline=-0.85ex]{\fill[fill opacity=0] (0,0) circle (0.85ex);}}

\date{}
\begin{document}
\maketitle

\begin{tikzpicture}[remember picture,overlay,shift={(current page.north west)}]
\node[anchor=north west,xshift=16.7cm,yshift=0.2cm]{\scalebox{0.48}[0.48]{\includegraphics[width=10cm]{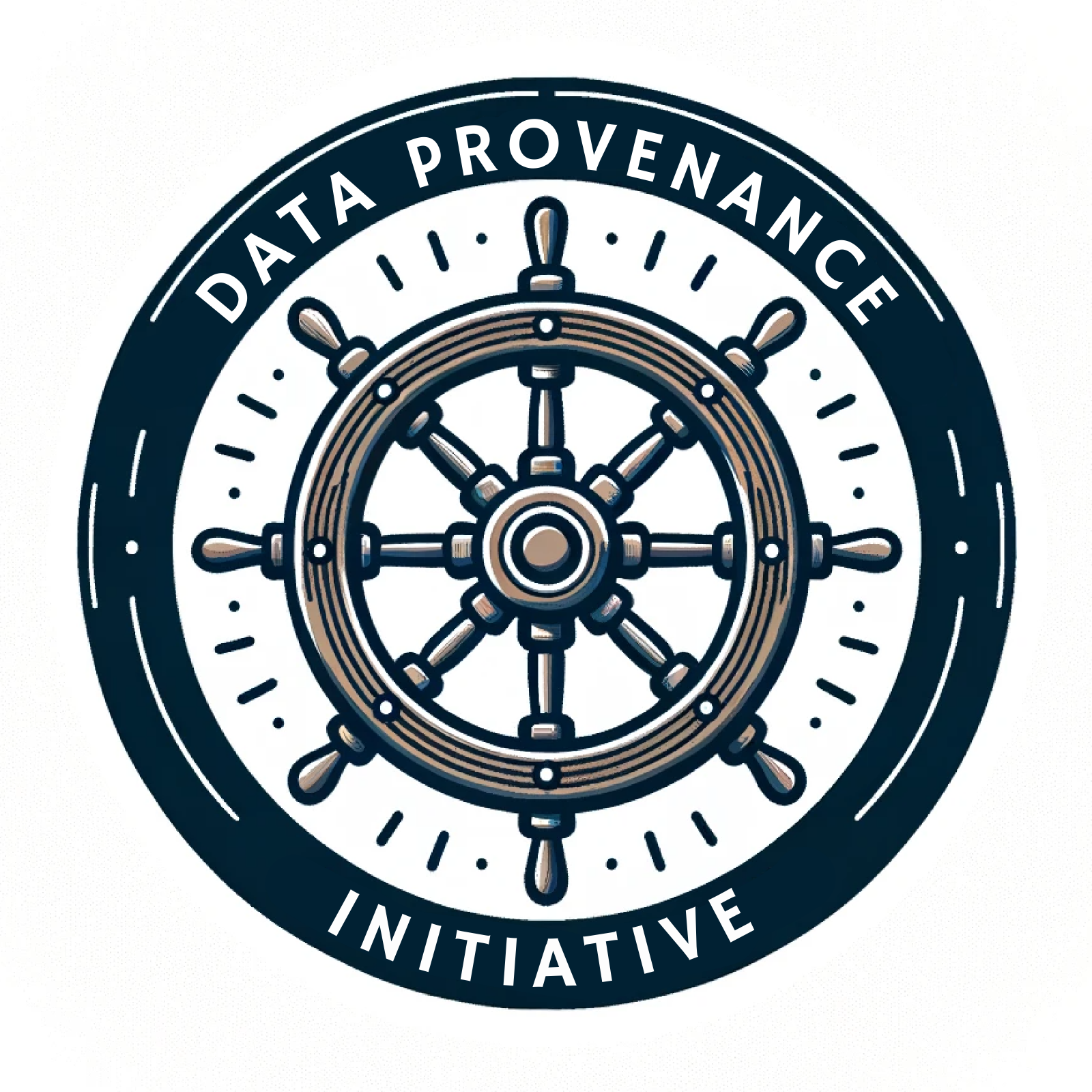}}};
\end{tikzpicture}


\vspace{-1mm}
\begin{abstract}
\noindent
The race to train language models on vast, diverse, and inconsistently documented datasets has raised pressing concerns about the legal and ethical risks for practitioners.
To remedy these practices threatening data transparency and understanding, we convene a multi-disciplinary effort between legal and machine learning experts to systematically audit and trace 1800+ text datasets. 
We develop tools and standards to trace the lineage of these datasets, from their source, creators, series of license conditions, properties, and subsequent use.
Our landscape analysis highlights the sharp divides in composition and focus of commercially open vs closed datasets, with closed datasets monopolizing important categories: lower resource languages, more creative tasks, richer topic variety, newer and more synthetic training data. 
This points to a deepening divide in the types of data that are made available under different license conditions, and heightened implications for jurisdictional legal interpretations of copyright and fair use.
We also observe frequent miscategorization of licenses on widely used dataset hosting sites, with license omission of 70\%+ and error rates of 50\%+.
This points to a crisis in misattribution and informed use of the most popular datasets driving many recent breakthroughs. 
As a contribution to ongoing improvements in dataset transparency and responsible use, we release our entire audit, with an interactive UI, the \textbf{Data Provenance Explorer}, which allows practitioners to trace and filter on data provenance for the most popular open source finetuning data collections: \url{www.dataprovenance.org}.
\end{abstract}

\vspace{-2mm}
\section{Introduction}
\label{introduction}
\vspace{-1mm}
The latest wave of language models, both public \citep{chung2022scaling,alpaca,koala_blogpost_2023} and proprietary~\citep{anil2023palm,openai2023gpt4,anthropic2023,yoo2022scalable} attribute their powerful abilities in large part to the diversity and richness of ever larger training datasets, including pre-training corpora, and finetuning datasets compiled by academics \citep{weifinetuned,sanh2021multitask, muennighoff2022crosslingual}, synthetically generated by models \citep{alpaca, selfinstruct2022}, or aggregated by platforms like Hugging Face \citep{lhoest2021datasets}.
Recent trends see practitioners combining and re-packaging thousands of datasets and web sources \citep{gao2020pile,refinedweb,wang2022benchmarking,longpre2023flan}, but despite some notable documentation efforts \citep{gaia, biderman2022datasheet}, there are diminishing efforts to attribute, document or understand the raw ingredients into new models \citep{dodge2021documenting,bandy2021addressing,bommasani2023foundation}.

\textbf{A Crisis in Data Transparency \& its Consequences.} Increasingly, widely used dataset collections are treated as monolithic, instead of a lineage of data sources, scraped (or model generated), curated, and annotated, often with multiple rounds of re-packaging (and re-licensing) by successive practitioners.
The disincentives to acknowledge this lineage stem both from the scale of modern data collection (the effort to properly attribute it), and the increased copyright scrutiny~\citep{tremblay2023openai}.
Together, these factors have seen fewer Datasheets~\citep{gebru2021datasheets}, non-disclosure of training sources~\citep{openai2023gpt4,anil2023palm,touvron2023llama}, and ultimately a decline in understanding training data \citep{sambasivan2021everyone,longpre2023pretrainers}.

This lack of understanding can lead to data leakages between training and test data \citep{elangovan-etal-2021-memorization, carlini2022quantifying}, expose personally identifiable information (PII) \citep{bubeck2023sparks}, present unintended biases or behaviours \citep{welbl2021challenges,xu2021detoxifying,pozzobon2023challenges}, and generally result in lower quality models than anticipated. Beyond these practical challenges, information gaps and documentation debt incur substantial ethical and legal risks.
For instance, model releases appear to contradict data terms of use (e.g., WizardCoder \citep{luo2023wizardcoder} licensed for commercial use, while training on commercially-prohibited OpenAI data), license revisions post-public release (with MPT-StoryTeller \citep{mosailml_jonathan_2023}), and even copyright lawsuits (e.g. Stability AI \citep{arstechnica} and OpenAI \citep{tremblay2023openai}).
As training models on data is both expensive and largely irreversible, these risks and challenges are not easily remedied. In this work, we term the combination of these indicators, including datasets' sourcing, creation and licensing heritage, as well as its characteristics, \textit{Data Provenance}.

\textbf{Unreliable Data Provenance \& Licensing.} Our work motivates the urgency of tooling that facilitates informed and responsible use of data in both pretraining and finetuning.
To empower practitioners to attribute data provenance, we develop a set of tools and standards to trace the data lineage of 44 of the most widely used and adopted text data collections, spanning 1800+ finetuning datasets.
We compile and expand relevant metadata with a much richer taxonomy than Hugging Face, Papers with Code, or other aggregators (see \cref{sec:data-attributes}).
With legal experts, we design a pipeline for tracing dataset provenance, including the original source of the dataset, the associated licenses, creators, and subsequent use.

As a byproduct of our work establishing the \textit{Data Provenance} of widely used datasets, we are able to characterize the AI data ecosystem/supply chain~\citep{cen2023aisupply,bommasani2023ecosystem}, as well as state of the field for policymakers, researchers and legal experts.
Our work points to a crisis in license laundering and informed usage of popular datasets, with systemic problems in sparse, ambiguous, or incorrect license documentation. 
Notably, we find that 70\%+ of licenses for popular datasets on GitHub and Hugging Face are ``Unspecified'', leaving a substantial information gap that is difficult to navigate in terms of legal responsibility. 
Second, the licenses that are attached to datasets uploaded to dataset sharing platforms are often inconsistent with the license ascribed by the original author of the dataset---our rigorous re-annotation of licenses finds that 66\% of analyzed Hugging Face licenses were in a different use category, often labeled as more permissive than the author's intended license.
As a result, much of this data is risky to use (or harmfully misleading) for practitioners who want to respect the data provenance of a work.
Our initiative reduces ``Unspecified`` licenses from 72\%+ to 30\% and attaches license URLs for under-resourced model developers to more confidently select appropriate data for their needs.
To this end, the Data Provenance Initiative supports attribution and responsible AI with the following contributions:
\vspace{-1mm}
\begin{enumerate}\itemsep0em
    \item \textbf{The most extensive known public audit of AI Data Provenance}, tracing the lineage of 1800+ text datasets (the ``\textbf{\dpic}''), their licenses, conditions, and sources.
    We demonstrate a growing adoption and reliance on software licenses in the AI community and synthesize observations into legal guidance for developers (\cref{sec:legal-discussion}).
    \item \textbf{The Data Provenance Explorer (\dpiex{})}\footnote{\url{www.dataprovenance.org}}, an open-source repository for downloading, filtering, and exploring data provenance and characteristics. 
    Our tools auto-generate \emph{Data Provenance Cards} for scalable symbolic attribution and future documentation best practices.
    \item \textbf{We find a sharp and widening divide between commercially open and closed data}, with the latter monopolizing more diverse and creative sources. We suggest a data collection focus to narrow this gap.
\end{enumerate}

\vspace{-3mm}
\section{The Initiative to Audit Data Provenance}

The Data Provenance Initiative's goal is to audit popular and widely used datasets with large-scale Legal and AI expert-guided annotation. 
We propose a base set of indicators necessary for tracing dataset lineage and understanding dataset risks (described in \cref{sec:data-attributes}).
As a first contribution of the initiative, we audit 44 instruction or ``alignment'' finetuning data collections composed of 1858 individual datasets, selected by experts for their widespread adoption and use in the community.
The selected collections and their variants see 100s to 10M+ monthly downloads on Hugging Face, with the datasets within these collections tallying to many more \cref{tab:collections}.

The initiative's initial focus on alignment finetuning datasets was decided based on their growing emphasis in the community for improving helpfulness, reducing harmfulness, and orienting models to human values~\citep{ouyang2022training}.
Some collections have overlapping datasets and examples, but we choose not to deduplicate to preserve the original design choices, that may include different templates, formatting, and filtering.
We remove datasets related to common benchmarks like MMLU~\citep{hendrycks2020measuring} and BigBench~\citep{srivastava2023beyond}.

\definecolor{id_color}{rgb}{0.8, 0.886, 0.768}
\definecolor{char_color}{rgb}{0.9137, 0.8588, 0.7019}
\definecolor{prov_color}{rgb}{0.8941, 0.7843, 0.8352}

\begin{figure*}[ht]
\centering
     \includegraphics[width=1.0\textwidth]{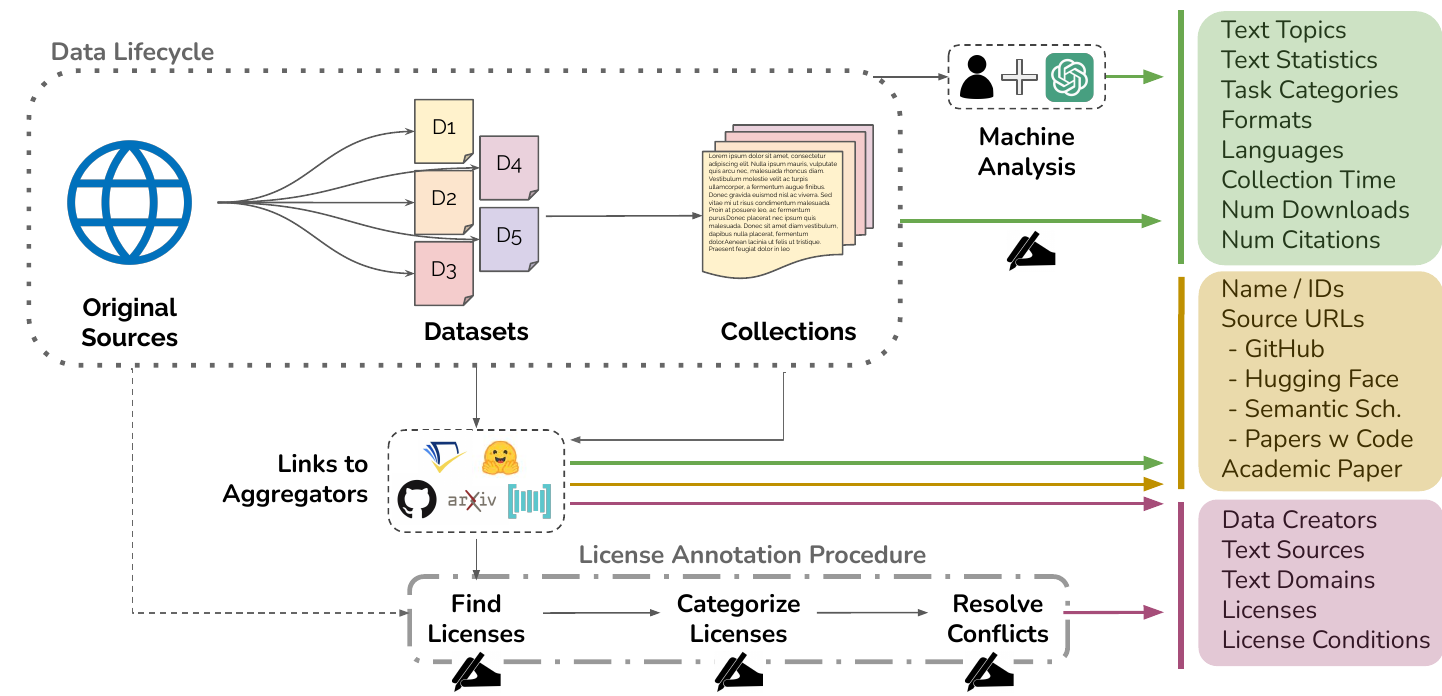}
      \caption{\textbf{The \dpic{} annotation pipeline uses human and human-assisted procedures to annotate dataset \colorbox{char_color}{Identifiers}, \colorbox{id_color}{Characteristics}, and \colorbox{prov_color}{Provenance}.}
      The \emph{Data Lifecycle} is traced, from the original sources (web scrapes, human or synthetic text), to curated datasets and packaged collections.
      Information is collected at each stage, not just the last.
      The \emph{License Annotation Procedure} is described in \cref{sec:license-collection}.}
       \label{data_collection}
       \vspace{-3mm}
\end{figure*}

\setlength{\tabcolsep}{1.9pt}
\definecolor{colorAnthropicHHPropertyCounts_Datasets}{RGB}{240,223,178}
\definecolor{colorDolly15kPropertyCounts_Datasets}{RGB}{245,239,222}
\definecolor{colorOpenAssistantPropertyCounts_Datasets}{RGB}{245,244,242}
\definecolor{colorFlanCollectionPropertyCounts_Datasets}{RGB}{182,227,220}
\definecolor{colorxP3xPropertyCounts_Datasets}{RGB}{179,226,219}
\definecolor{colorTasksourceIns.PropertyCounts_Datasets}{RGB}{196,232,227}
\definecolor{colorLAIONOIGPropertyCounts_Datasets}{RGB}{242,244,244}
\definecolor{colorSHPPropertyCounts_Datasets}{RGB}{245,244,242}
\definecolor{colorShareGPTPropertyCounts_Datasets}{RGB}{240,223,178}
\definecolor{colorSelf-InstructPropertyCounts_Datasets}{RGB}{240,223,178}
\definecolor{colorWebGPTPropertyCounts_Datasets}{RGB}{245,237,214}
\definecolor{colorOpenAISumm.PropertyCounts_Datasets}{RGB}{240,223,178}
\definecolor{colorAiroborosPropertyCounts_Datasets}{RGB}{240,223,178}
\definecolor{colorAlpacaPropertyCounts_Datasets}{RGB}{240,223,178}
\definecolor{colorBaizeChatPropertyCounts_Datasets}{RGB}{245,236,210}
\definecolor{colorBookSumPropertyCounts_Datasets}{RGB}{240,223,178}
\definecolor{colorCamelAISci.PropertyCounts_Datasets}{RGB}{245,234,204}
\definecolor{colorCoTColl.PropertyCounts_Datasets}{RGB}{245,238,218}
\definecolor{colorCodeAlpacaPropertyCounts_Datasets}{RGB}{240,223,178}
\definecolor{colorGPT-4-AlpacaPropertyCounts_Datasets}{RGB}{240,223,178}
\definecolor{colorGPTeacherPropertyCounts_Datasets}{RGB}{245,236,210}
\definecolor{colorGorillaPropertyCounts_Datasets}{RGB}{240,223,178}
\definecolor{colorHC3PropertyCounts_Datasets}{RGB}{245,241,232}
\definecolor{colorJokeExpl.PropertyCounts_Datasets}{RGB}{240,223,178}
\definecolor{colorLIMAPropertyCounts_Datasets}{RGB}{245,237,214}
\definecolor{colorLongformPropertyCounts_Datasets}{RGB}{245,239,222}
\definecolor{colorGPT4AllJPropertyCounts_Datasets}{RGB}{245,239,222}
\definecolor{colorOpenOrcaPropertyCounts_Datasets}{RGB}{245,236,210}
\definecolor{colorTool-LlamaPropertyCounts_Datasets}{RGB}{240,223,178}
\definecolor{colorUltraChatPropertyCounts_Datasets}{RGB}{240,223,178}
\definecolor{colorUnnaturalInstr.PropertyCounts_Datasets}{RGB}{240,223,178}
\definecolor{colorEvol-Instr.PropertyCounts_Datasets}{RGB}{245,232,196}
\definecolor{colorStarCoderPropertyCounts_Datasets}{RGB}{240,223,178}
\definecolor{colorTinyStoriesPropertyCounts_Datasets}{RGB}{240,223,178}
\definecolor{colorStackExchangePropertyCounts_Datasets}{RGB}{240,223,178}
\definecolor{colorTasksourceSTPropertyCounts_Datasets}{RGB}{200,234,229}
\definecolor{colorCommitPackFTPropertyCounts_Datasets}{RGB}{196,232,227}
\definecolor{colorOpAsstOctoPackPropertyCounts_Datasets}{RGB}{240,223,178}
\definecolor{colorAnthropicHHPropertyCounts_Dialogs}{RGB}{245,242,234}
\definecolor{colorDolly15kPropertyCounts_Dialogs}{RGB}{245,237,214}
\definecolor{colorOpenAssistantPropertyCounts_Dialogs}{RGB}{245,236,210}
\definecolor{colorFlanCollectionPropertyCounts_Dialogs}{RGB}{222,239,237}
\definecolor{colorxP3xPropertyCounts_Dialogs}{RGB}{179,226,219}
\definecolor{colorTasksourceIns.PropertyCounts_Dialogs}{RGB}{231,241,240}
\definecolor{colorLAIONOIGPropertyCounts_Dialogs}{RGB}{222,239,237}
\definecolor{colorSHPPropertyCounts_Dialogs}{RGB}{245,243,240}
\definecolor{colorShareGPTPropertyCounts_Dialogs}{RGB}{245,240,228}
\definecolor{colorSelf-InstructPropertyCounts_Dialogs}{RGB}{245,240,228}
\definecolor{colorWebGPTPropertyCounts_Dialogs}{RGB}{245,237,216}
\definecolor{colorOpenAISumm.PropertyCounts_Dialogs}{RGB}{245,241,230}
\definecolor{colorAiroborosPropertyCounts_Dialogs}{RGB}{245,237,214}
\definecolor{colorAlpacaPropertyCounts_Dialogs}{RGB}{245,239,224}
\definecolor{colorBaizeChatPropertyCounts_Dialogs}{RGB}{245,242,236}
\definecolor{colorBookSumPropertyCounts_Dialogs}{RGB}{245,235,206}
\definecolor{colorCamelAISci.PropertyCounts_Dialogs}{RGB}{245,240,226}
\definecolor{colorCoTColl.PropertyCounts_Dialogs}{RGB}{235,242,241}
\definecolor{colorCodeAlpacaPropertyCounts_Dialogs}{RGB}{245,237,216}
\definecolor{colorGPT-4-AlpacaPropertyCounts_Dialogs}{RGB}{245,239,224}
\definecolor{colorGPTeacherPropertyCounts_Dialogs}{RGB}{245,241,230}
\definecolor{colorGorillaPropertyCounts_Dialogs}{RGB}{245,237,214}
\definecolor{colorHC3PropertyCounts_Dialogs}{RGB}{245,239,222}
\definecolor{colorJokeExpl.PropertyCounts_Dialogs}{RGB}{239,221,175}
\definecolor{colorLIMAPropertyCounts_Dialogs}{RGB}{244,229,189}
\definecolor{colorLongformPropertyCounts_Dialogs}{RGB}{245,237,216}
\definecolor{colorGPT4AllJPropertyCounts_Dialogs}{RGB}{242,244,244}
\definecolor{colorOpenOrcaPropertyCounts_Dialogs}{RGB}{229,241,239}
\definecolor{colorTool-LlamaPropertyCounts_Dialogs}{RGB}{245,239,222}
\definecolor{colorUltraChatPropertyCounts_Dialogs}{RGB}{236,243,242}
\definecolor{colorUnnaturalInstr.PropertyCounts_Dialogs}{RGB}{245,240,226}
\definecolor{colorEvol-Instr.PropertyCounts_Dialogs}{RGB}{245,242,236}
\definecolor{colorStarCoderPropertyCounts_Dialogs}{RGB}{243,227,186}
\definecolor{colorTinyStoriesPropertyCounts_Dialogs}{RGB}{245,236,212}
\definecolor{colorStackExchangePropertyCounts_Dialogs}{RGB}{222,239,237}
\definecolor{colorTasksourceSTPropertyCounts_Dialogs}{RGB}{245,243,240}
\definecolor{colorCommitPackFTPropertyCounts_Dialogs}{RGB}{244,244,244}
\definecolor{colorOpAsstOctoPackPropertyCounts_Dialogs}{RGB}{245,236,210}
\definecolor{colorAnthropicHHPropertyCounts_Tasks}{RGB}{245,240,228}
\definecolor{colorDolly15kPropertyCounts_Tasks}{RGB}{240,243,243}
\definecolor{colorOpenAssistantPropertyCounts_Tasks}{RGB}{245,244,242}
\definecolor{colorFlanCollectionPropertyCounts_Tasks}{RGB}{179,226,219}
\definecolor{colorxP3xPropertyCounts_Tasks}{RGB}{240,243,243}
\definecolor{colorTasksourceIns.PropertyCounts_Tasks}{RGB}{202,234,230}
\definecolor{colorLAIONOIGPropertyCounts_Tasks}{RGB}{204,235,230}
\definecolor{colorSHPPropertyCounts_Tasks}{RGB}{233,242,240}
\definecolor{colorShareGPTPropertyCounts_Tasks}{RGB}{217,238,235}
\definecolor{colorSelf-InstructPropertyCounts_Tasks}{RGB}{233,242,240}
\definecolor{colorWebGPTPropertyCounts_Tasks}{RGB}{245,244,242}
\definecolor{colorOpenAISumm.PropertyCounts_Tasks}{RGB}{240,243,243}
\definecolor{colorAiroborosPropertyCounts_Tasks}{RGB}{240,243,243}
\definecolor{colorAlpacaPropertyCounts_Tasks}{RGB}{220,239,236}
\definecolor{colorBaizeChatPropertyCounts_Tasks}{RGB}{204,235,230}
\definecolor{colorBookSumPropertyCounts_Tasks}{RGB}{245,244,242}
\definecolor{colorCamelAISci.PropertyCounts_Tasks}{RGB}{245,236,212}
\definecolor{colorCoTColl.PropertyCounts_Tasks}{RGB}{204,235,230}
\definecolor{colorCodeAlpacaPropertyCounts_Tasks}{RGB}{245,240,228}
\definecolor{colorGPT-4-AlpacaPropertyCounts_Tasks}{RGB}{226,240,238}
\definecolor{colorGPTeacherPropertyCounts_Tasks}{RGB}{220,239,236}
\definecolor{colorGorillaPropertyCounts_Tasks}{RGB}{245,244,242}
\definecolor{colorHC3PropertyCounts_Tasks}{RGB}{233,242,240}
\definecolor{colorJokeExpl.PropertyCounts_Tasks}{RGB}{245,236,212}
\definecolor{colorLIMAPropertyCounts_Tasks}{RGB}{211,237,233}
\definecolor{colorLongformPropertyCounts_Tasks}{RGB}{208,236,232}
\definecolor{colorGPT4AllJPropertyCounts_Tasks}{RGB}{211,237,233}
\definecolor{colorOpenOrcaPropertyCounts_Tasks}{RGB}{208,236,232}
\definecolor{colorTool-LlamaPropertyCounts_Tasks}{RGB}{245,236,212}
\definecolor{colorUltraChatPropertyCounts_Tasks}{RGB}{226,240,238}
\definecolor{colorUnnaturalInstr.PropertyCounts_Tasks}{RGB}{245,244,242}
\definecolor{colorEvol-Instr.PropertyCounts_Tasks}{RGB}{208,236,232}
\definecolor{colorStarCoderPropertyCounts_Tasks}{RGB}{239,221,175}
\definecolor{colorTinyStoriesPropertyCounts_Tasks}{RGB}{245,244,242}
\definecolor{colorStackExchangePropertyCounts_Tasks}{RGB}{239,221,175}
\definecolor{colorTasksourceSTPropertyCounts_Tasks}{RGB}{196,232,227}
\definecolor{colorCommitPackFTPropertyCounts_Tasks}{RGB}{239,221,175}
\definecolor{colorOpAsstOctoPackPropertyCounts_Tasks}{RGB}{245,240,228}
\definecolor{colorAnthropicHHPropertyCounts_Langs}{RGB}{240,223,178}
\definecolor{colorDolly15kPropertyCounts_Langs}{RGB}{240,223,178}
\definecolor{colorOpenAssistantPropertyCounts_Langs}{RGB}{240,243,243}
\definecolor{colorFlanCollectionPropertyCounts_Langs}{RGB}{227,240,239}
\definecolor{colorxP3xPropertyCounts_Langs}{RGB}{185,228,221}
\definecolor{colorTasksourceIns.PropertyCounts_Langs}{RGB}{240,223,178}
\definecolor{colorLAIONOIGPropertyCounts_Langs}{RGB}{240,223,178}
\definecolor{colorSHPPropertyCounts_Langs}{RGB}{245,232,196}
\definecolor{colorShareGPTPropertyCounts_Langs}{RGB}{240,223,178}
\definecolor{colorSelf-InstructPropertyCounts_Langs}{RGB}{245,232,196}
\definecolor{colorWebGPTPropertyCounts_Langs}{RGB}{240,223,178}
\definecolor{colorOpenAISumm.PropertyCounts_Langs}{RGB}{240,223,178}
\definecolor{colorAiroborosPropertyCounts_Langs}{RGB}{245,232,196}
\definecolor{colorAlpacaPropertyCounts_Langs}{RGB}{240,223,178}
\definecolor{colorBaizeChatPropertyCounts_Langs}{RGB}{245,232,196}
\definecolor{colorBookSumPropertyCounts_Langs}{RGB}{240,223,178}
\definecolor{colorCamelAISci.PropertyCounts_Langs}{RGB}{240,223,178}
\definecolor{colorCoTColl.PropertyCounts_Langs}{RGB}{245,240,226}
\definecolor{colorCodeAlpacaPropertyCounts_Langs}{RGB}{245,232,196}
\definecolor{colorGPT-4-AlpacaPropertyCounts_Langs}{RGB}{240,223,178}
\definecolor{colorGPTeacherPropertyCounts_Langs}{RGB}{245,232,196}
\definecolor{colorGorillaPropertyCounts_Langs}{RGB}{245,232,196}
\definecolor{colorHC3PropertyCounts_Langs}{RGB}{245,232,196}
\definecolor{colorJokeExpl.PropertyCounts_Langs}{RGB}{240,223,178}
\definecolor{colorLIMAPropertyCounts_Langs}{RGB}{245,232,196}
\definecolor{colorLongformPropertyCounts_Langs}{RGB}{240,223,178}
\definecolor{colorGPT4AllJPropertyCounts_Langs}{RGB}{240,223,178}
\definecolor{colorOpenOrcaPropertyCounts_Langs}{RGB}{240,223,178}
\definecolor{colorTool-LlamaPropertyCounts_Langs}{RGB}{245,232,196}
\definecolor{colorUltraChatPropertyCounts_Langs}{RGB}{240,223,178}
\definecolor{colorUnnaturalInstr.PropertyCounts_Langs}{RGB}{240,223,178}
\definecolor{colorEvol-Instr.PropertyCounts_Langs}{RGB}{245,232,196}
\definecolor{colorStarCoderPropertyCounts_Langs}{RGB}{245,232,196}
\definecolor{colorTinyStoriesPropertyCounts_Langs}{RGB}{240,223,178}
\definecolor{colorStackExchangePropertyCounts_Langs}{RGB}{245,232,196}
\definecolor{colorTasksourceSTPropertyCounts_Langs}{RGB}{240,223,178}
\definecolor{colorCommitPackFTPropertyCounts_Langs}{RGB}{179,226,219}
\definecolor{colorOpAsstOctoPackPropertyCounts_Langs}{RGB}{240,243,243}
\definecolor{colorAnthropicHHPropertyCounts_Topics}{RGB}{240,223,178}
\definecolor{colorDolly15kPropertyCounts_Topics}{RGB}{245,238,218}
\definecolor{colorOpenAssistantPropertyCounts_Topics}{RGB}{245,244,244}
\definecolor{colorFlanCollectionPropertyCounts_Topics}{RGB}{179,226,219}
\definecolor{colorxP3xPropertyCounts_Topics}{RGB}{235,242,241}
\definecolor{colorTasksourceIns.PropertyCounts_Topics}{RGB}{202,234,230}
\definecolor{colorLAIONOIGPropertyCounts_Topics}{RGB}{233,242,240}
\definecolor{colorSHPPropertyCounts_Topics}{RGB}{235,242,241}
\definecolor{colorShareGPTPropertyCounts_Topics}{RGB}{240,223,178}
\definecolor{colorSelf-InstructPropertyCounts_Topics}{RGB}{240,223,178}
\definecolor{colorWebGPTPropertyCounts_Topics}{RGB}{245,237,216}
\definecolor{colorOpenAISumm.PropertyCounts_Topics}{RGB}{240,223,178}
\definecolor{colorAiroborosPropertyCounts_Topics}{RGB}{240,223,178}
\definecolor{colorAlpacaPropertyCounts_Topics}{RGB}{240,223,178}
\definecolor{colorBaizeChatPropertyCounts_Topics}{RGB}{245,237,216}
\definecolor{colorBookSumPropertyCounts_Topics}{RGB}{240,223,178}
\definecolor{colorCamelAISci.PropertyCounts_Topics}{RGB}{245,236,210}
\definecolor{colorCoTColl.PropertyCounts_Topics}{RGB}{245,236,210}
\definecolor{colorCodeAlpacaPropertyCounts_Topics}{RGB}{240,223,178}
\definecolor{colorGPT-4-AlpacaPropertyCounts_Topics}{RGB}{240,223,178}
\definecolor{colorGPTeacherPropertyCounts_Topics}{RGB}{245,237,214}
\definecolor{colorGorillaPropertyCounts_Topics}{RGB}{240,223,178}
\definecolor{colorHC3PropertyCounts_Topics}{RGB}{245,244,244}
\definecolor{colorJokeExpl.PropertyCounts_Topics}{RGB}{240,223,178}
\definecolor{colorLIMAPropertyCounts_Topics}{RGB}{245,238,220}
\definecolor{colorLongformPropertyCounts_Topics}{RGB}{245,241,232}
\definecolor{colorGPT4AllJPropertyCounts_Topics}{RGB}{245,240,228}
\definecolor{colorOpenOrcaPropertyCounts_Topics}{RGB}{245,236,210}
\definecolor{colorTool-LlamaPropertyCounts_Topics}{RGB}{240,223,178}
\definecolor{colorUltraChatPropertyCounts_Topics}{RGB}{241,224,181}
\definecolor{colorUnnaturalInstr.PropertyCounts_Topics}{RGB}{240,223,178}
\definecolor{colorEvol-Instr.PropertyCounts_Topics}{RGB}{245,232,196}
\definecolor{colorStarCoderPropertyCounts_Topics}{RGB}{240,223,178}
\definecolor{colorTinyStoriesPropertyCounts_Topics}{RGB}{240,223,178}
\definecolor{colorStackExchangePropertyCounts_Topics}{RGB}{240,223,178}
\definecolor{colorTasksourceSTPropertyCounts_Topics}{RGB}{208,236,232}
\definecolor{colorCommitPackFTPropertyCounts_Topics}{RGB}{196,232,227}
\definecolor{colorOpAsstOctoPackPropertyCounts_Topics}{RGB}{240,223,178}
\definecolor{colorAnthropicHHPropertyCounts_Domains}{RGB}{240,223,178}
\definecolor{colorDolly15kPropertyCounts_Domains}{RGB}{240,223,178}
\definecolor{colorOpenAssistantPropertyCounts_Domains}{RGB}{240,223,178}
\definecolor{colorFlanCollectionPropertyCounts_Domains}{RGB}{179,226,219}
\definecolor{colorxP3xPropertyCounts_Domains}{RGB}{206,235,231}
\definecolor{colorTasksourceIns.PropertyCounts_Domains}{RGB}{187,229,223}
\definecolor{colorLAIONOIGPropertyCounts_Domains}{RGB}{215,237,234}
\definecolor{colorSHPPropertyCounts_Domains}{RGB}{240,223,178}
\definecolor{colorShareGPTPropertyCounts_Domains}{RGB}{245,236,210}
\definecolor{colorSelf-InstructPropertyCounts_Domains}{RGB}{240,223,178}
\definecolor{colorWebGPTPropertyCounts_Domains}{RGB}{245,240,226}
\definecolor{colorOpenAISumm.PropertyCounts_Domains}{RGB}{240,223,178}
\definecolor{colorAiroborosPropertyCounts_Domains}{RGB}{240,223,178}
\definecolor{colorAlpacaPropertyCounts_Domains}{RGB}{240,223,178}
\definecolor{colorBaizeChatPropertyCounts_Domains}{RGB}{245,240,226}
\definecolor{colorBookSumPropertyCounts_Domains}{RGB}{240,223,178}
\definecolor{colorCamelAISci.PropertyCounts_Domains}{RGB}{240,223,178}
\definecolor{colorCoTColl.PropertyCounts_Domains}{RGB}{240,223,178}
\definecolor{colorCodeAlpacaPropertyCounts_Domains}{RGB}{240,223,178}
\definecolor{colorGPT-4-AlpacaPropertyCounts_Domains}{RGB}{240,223,178}
\definecolor{colorGPTeacherPropertyCounts_Domains}{RGB}{240,223,178}
\definecolor{colorGorillaPropertyCounts_Domains}{RGB}{245,236,210}
\definecolor{colorHC3PropertyCounts_Domains}{RGB}{236,243,242}
\definecolor{colorJokeExpl.PropertyCounts_Domains}{RGB}{240,223,178}
\definecolor{colorLIMAPropertyCounts_Domains}{RGB}{236,243,242}
\definecolor{colorLongformPropertyCounts_Domains}{RGB}{245,243,238}
\definecolor{colorGPT4AllJPropertyCounts_Domains}{RGB}{240,223,178}
\definecolor{colorOpenOrcaPropertyCounts_Domains}{RGB}{179,226,219}
\definecolor{colorTool-LlamaPropertyCounts_Domains}{RGB}{240,223,178}
\definecolor{colorUltraChatPropertyCounts_Domains}{RGB}{245,236,210}
\definecolor{colorUnnaturalInstr.PropertyCounts_Domains}{RGB}{240,223,178}
\definecolor{colorEvol-Instr.PropertyCounts_Domains}{RGB}{240,223,178}
\definecolor{colorStarCoderPropertyCounts_Domains}{RGB}{240,223,178}
\definecolor{colorTinyStoriesPropertyCounts_Domains}{RGB}{240,223,178}
\definecolor{colorStackExchangePropertyCounts_Domains}{RGB}{240,223,178}
\definecolor{colorTasksourceSTPropertyCounts_Domains}{RGB}{196,232,227}
\definecolor{colorCommitPackFTPropertyCounts_Domains}{RGB}{240,223,178}
\definecolor{colorOpAsstOctoPackPropertyCounts_Domains}{RGB}{240,223,178}
\definecolor{colorAnthropicHHPropertyCounts_Downs}{RGB}{206,235,231}
\definecolor{colorDolly15kPropertyCounts_Downs}{RGB}{179,226,219}
\definecolor{colorOpenAssistantPropertyCounts_Downs}{RGB}{213,237,234}
\definecolor{colorFlanCollectionPropertyCounts_Downs}{RGB}{211,237,233}
\definecolor{colorxP3xPropertyCounts_Downs}{RGB}{222,239,237}
\definecolor{colorTasksourceIns.PropertyCounts_Downs}{RGB}{240,243,243}
\definecolor{colorLAIONOIGPropertyCounts_Downs}{RGB}{226,240,238}
\definecolor{colorSHPPropertyCounts_Downs}{RGB}{217,238,235}
\definecolor{colorShareGPTPropertyCounts_Downs}{RGB}{224,240,237}
\definecolor{colorSelf-InstructPropertyCounts_Downs}{RGB}{218,238,235}
\definecolor{colorWebGPTPropertyCounts_Downs}{RGB}{222,239,237}
\definecolor{colorOpenAISumm.PropertyCounts_Downs}{RGB}{213,237,234}
\definecolor{colorAiroborosPropertyCounts_Downs}{RGB}{222,239,237}
\definecolor{colorAlpacaPropertyCounts_Downs}{RGB}{204,235,230}
\definecolor{colorBaizeChatPropertyCounts_Downs}{RGB}{224,240,237}
\definecolor{colorBookSumPropertyCounts_Downs}{RGB}{224,240,237}
\definecolor{colorCamelAISci.PropertyCounts_Downs}{RGB}{233,242,240}
\definecolor{colorCoTColl.PropertyCounts_Downs}{RGB}{226,240,238}
\definecolor{colorCodeAlpacaPropertyCounts_Downs}{RGB}{217,238,235}
\definecolor{colorGPT-4-AlpacaPropertyCounts_Downs}{RGB}{222,239,237}
\definecolor{colorGPTeacherPropertyCounts_Downs}{RGB}{231,241,240}
\definecolor{colorGorillaPropertyCounts_Downs}{RGB}{231,241,240}
\definecolor{colorHC3PropertyCounts_Downs}{RGB}{220,239,236}
\definecolor{colorJokeExpl.PropertyCounts_Downs}{RGB}{235,242,241}
\definecolor{colorLIMAPropertyCounts_Downs}{RGB}{218,238,235}
\definecolor{colorLongformPropertyCounts_Downs}{RGB}{218,238,235}
\definecolor{colorGPT4AllJPropertyCounts_Downs}{RGB}{224,240,237}
\definecolor{colorOpenOrcaPropertyCounts_Downs}{RGB}{209,236,232}
\definecolor{colorTool-LlamaPropertyCounts_Downs}{RGB}{240,223,178}
\definecolor{colorUltraChatPropertyCounts_Downs}{RGB}{220,239,236}
\definecolor{colorUnnaturalInstr.PropertyCounts_Downs}{RGB}{238,243,242}
\definecolor{colorEvol-Instr.PropertyCounts_Downs}{RGB}{220,239,236}
\definecolor{colorStarCoderPropertyCounts_Downs}{RGB}{227,240,239}
\definecolor{colorTinyStoriesPropertyCounts_Downs}{RGB}{213,237,234}
\definecolor{colorStackExchangePropertyCounts_Downs}{RGB}{224,240,237}
\definecolor{colorTasksourceSTPropertyCounts_Downs}{RGB}{229,241,239}
\definecolor{colorCommitPackFTPropertyCounts_Downs}{RGB}{217,238,235}
\definecolor{colorOpAsstOctoPackPropertyCounts_Downs}{RGB}{235,242,241}
\definecolor{colorAnthropicHHTextLens_Inpt}{RGB}{240,223,178}
\definecolor{colorDolly15kTextLens_Inpt}{RGB}{245,239,224}
\definecolor{colorOpenAssistantTextLens_Inpt}{RGB}{246,232,195}
\definecolor{colorFlanCollectionTextLens_Inpt}{RGB}{233,242,240}
\definecolor{colorxP3xTextLens_Inpt}{RGB}{245,241,232}
\definecolor{colorTasksourceIns.TextLens_Inpt}{RGB}{245,241,230}
\definecolor{colorLAIONOIGTextLens_Inpt}{RGB}{245,238,220}
\definecolor{colorSHPTextLens_Inpt}{RGB}{245,244,242}
\definecolor{colorShareGPTTextLens_Inpt}{RGB}{245,237,216}
\definecolor{colorSelf-InstructTextLens_Inpt}{RGB}{245,233,198}
\definecolor{colorWebGPTTextLens_Inpt}{RGB}{245,243,238}
\definecolor{colorOpenAISumm.TextLens_Inpt}{RGB}{238,243,242}
\definecolor{colorAiroborosTextLens_Inpt}{RGB}{245,238,220}
\definecolor{colorAlpacaTextLens_Inpt}{RGB}{245,241,230}
\definecolor{colorBaizeChatTextLens_Inpt}{RGB}{240,223,178}
\definecolor{colorBookSumTextLens_Inpt}{RGB}{179,226,219}
\definecolor{colorCamelAISci.TextLens_Inpt}{RGB}{245,235,206}
\definecolor{colorCoTColl.TextLens_Inpt}{RGB}{245,243,238}
\definecolor{colorCodeAlpacaTextLens_Inpt}{RGB}{244,229,189}
\definecolor{colorGPT-4-AlpacaTextLens_Inpt}{RGB}{245,232,196}
\definecolor{colorGPTeacherTextLens_Inpt}{RGB}{245,236,210}
\definecolor{colorGorillaTextLens_Inpt}{RGB}{246,232,195}
\definecolor{colorHC3TextLens_Inpt}{RGB}{246,232,195}
\definecolor{colorJokeExpl.TextLens_Inpt}{RGB}{243,227,186}
\definecolor{colorLIMATextLens_Inpt}{RGB}{245,236,210}
\definecolor{colorLongformTextLens_Inpt}{RGB}{245,243,240}
\definecolor{colorGPT4AllJTextLens_Inpt}{RGB}{245,244,242}
\definecolor{colorOpenOrcaTextLens_Inpt}{RGB}{240,243,243}
\definecolor{colorTool-LlamaTextLens_Inpt}{RGB}{200,234,229}
\definecolor{colorUltraChatTextLens_Inpt}{RGB}{245,237,216}
\definecolor{colorUnnaturalInstr.TextLens_Inpt}{RGB}{245,238,218}
\definecolor{colorEvol-Instr.TextLens_Inpt}{RGB}{245,241,232}
\definecolor{colorStarCoderTextLens_Inpt}{RGB}{245,235,206}
\definecolor{colorTinyStoriesTextLens_Inpt}{RGB}{245,241,230}
\definecolor{colorStackExchangeTextLens_Inpt}{RGB}{240,243,243}
\definecolor{colorTasksourceSTTextLens_Inpt}{RGB}{224,240,237}
\definecolor{colorCommitPackFTTextLens_Inpt}{RGB}{245,242,236}
\definecolor{colorOpAsstOctoPackTextLens_Inpt}{RGB}{246,232,195}
\definecolor{colorAnthropicHHTextLens_Tgt}{RGB}{245,241,230}
\definecolor{colorDolly15kTextLens_Tgt}{RGB}{245,241,232}
\definecolor{colorOpenAssistantTextLens_Tgt}{RGB}{245,243,240}
\definecolor{colorFlanCollectionTextLens_Tgt}{RGB}{245,238,218}
\definecolor{colorxP3xTextLens_Tgt}{RGB}{245,242,234}
\definecolor{colorTasksourceIns.TextLens_Tgt}{RGB}{246,232,195}
\definecolor{colorLAIONOIGTextLens_Tgt}{RGB}{245,243,238}
\definecolor{colorSHPTextLens_Tgt}{RGB}{245,242,236}
\definecolor{colorShareGPTTextLens_Tgt}{RGB}{244,244,244}
\definecolor{colorSelf-InstructTextLens_Tgt}{RGB}{245,237,216}
\definecolor{colorWebGPTTextLens_Tgt}{RGB}{245,243,240}
\definecolor{colorOpenAISumm.TextLens_Tgt}{RGB}{245,238,220}
\definecolor{colorAiroborosTextLens_Tgt}{RGB}{242,244,244}
\definecolor{colorAlpacaTextLens_Tgt}{RGB}{245,240,228}
\definecolor{colorBaizeChatTextLens_Tgt}{RGB}{245,240,226}
\definecolor{colorBookSumTextLens_Tgt}{RGB}{240,243,243}
\definecolor{colorCamelAISci.TextLens_Tgt}{RGB}{236,243,242}
\definecolor{colorCoTColl.TextLens_Tgt}{RGB}{245,240,228}
\definecolor{colorCodeAlpacaTextLens_Tgt}{RGB}{245,239,224}
\definecolor{colorGPT-4-AlpacaTextLens_Tgt}{RGB}{245,242,236}
\definecolor{colorGPTeacherTextLens_Tgt}{RGB}{245,241,232}
\definecolor{colorGorillaTextLens_Tgt}{RGB}{245,236,212}
\definecolor{colorHC3TextLens_Tgt}{RGB}{245,243,238}
\definecolor{colorJokeExpl.TextLens_Tgt}{RGB}{245,242,236}
\definecolor{colorLIMATextLens_Tgt}{RGB}{235,242,241}
\definecolor{colorLongformTextLens_Tgt}{RGB}{238,243,242}
\definecolor{colorGPT4AllJTextLens_Tgt}{RGB}{242,244,244}
\definecolor{colorOpenOrcaTextLens_Tgt}{RGB}{245,242,236}
\definecolor{colorTool-LlamaTextLens_Tgt}{RGB}{244,244,244}
\definecolor{colorUltraChatTextLens_Tgt}{RGB}{242,244,244}
\definecolor{colorUnnaturalInstr.TextLens_Tgt}{RGB}{245,236,210}
\definecolor{colorEvol-Instr.TextLens_Tgt}{RGB}{240,243,243}
\definecolor{colorStarCoderTextLens_Tgt}{RGB}{245,242,236}
\definecolor{colorTinyStoriesTextLens_Tgt}{RGB}{179,226,219}
\definecolor{colorStackExchangeTextLens_Tgt}{RGB}{245,244,242}
\definecolor{colorTasksourceSTTextLens_Tgt}{RGB}{240,223,178}
\definecolor{colorCommitPackFTTextLens_Tgt}{RGB}{245,244,242}
\definecolor{colorOpAsstOctoPackTextLens_Tgt}{RGB}{245,244,242}

\begin{table*}
\centering
\resizebox{\textwidth}{!}{
\begin{tabular}{l|ccccccc|rr|cp{0.3cm}p{0.3cm}p{0.3cm}p{0.3cm}p{0.3cm}cp{0.3cm}}
\toprule
 \textsc{Collection} & \multicolumn{7}{c}{\textsc{Property Counts}} & \multicolumn{2}{c}{\textsc{Text Lens}} & \multicolumn{8}{c}{\textsc{Dataset Types}} \\
 & \textsc{\thead{Datasets}} & \textsc{\thead{Dialogs}} & \textsc{\thead{Tasks}} & \textsc{\thead{Langs}} & \textsc{\thead{Topics}} & \textsc{\thead{Domains}} & \textsc{\thead{Downs}} & \textsc{\thead{Inpt}} & \textsc{\thead{Tgt}} & \textsc{\thead{Source}} & \textsc{\thead{Z}} & \textsc{\thead{F}} & \textsc{\thead{C}} & \textsc{\thead{R}} & \textsc{\thead{M}} & \textsc{\thead{Use}} & \textsc{\thead{O}} \\
\midrule
Airoboros & \cellcolor{colorOpAsstOctoPackPropertyCounts_Datasets}{1} & \cellcolor{colorAiroborosPropertyCounts_Dialogs}{17k} & \cellcolor{colorAiroborosPropertyCounts_Tasks}{5} & \cellcolor{colorStackExchangePropertyCounts_Langs}{2} & \cellcolor{colorOpAsstOctoPackPropertyCounts_Topics}{10} & \cellcolor{colorOpAsstOctoPackPropertyCounts_Domains}{1} & \cellcolor{colorAiroborosPropertyCounts_Downs}{1k} & \cellcolor{colorAiroborosTextLens_Inpt}{347} & \cellcolor{colorAiroborosTextLens_Tgt}{1k} & \emojiblank\emoji{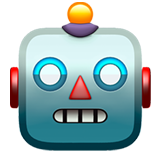} & \greencheck & \emojiblank & \emojiblank & \emojiblank & \emojiblank & \TransparentCircle \TransparentCircle \NCDataCircle & \greencheck \\
Alpaca & \cellcolor{colorOpAsstOctoPackPropertyCounts_Datasets}{1} & \cellcolor{colorAlpacaPropertyCounts_Dialogs}{52k} & \cellcolor{colorGPTeacherPropertyCounts_Tasks}{8} & \cellcolor{colorTasksourceSTPropertyCounts_Langs}{1} & \cellcolor{colorOpAsstOctoPackPropertyCounts_Topics}{10} & \cellcolor{colorOpAsstOctoPackPropertyCounts_Domains}{1} & \cellcolor{colorAlpacaPropertyCounts_Downs}{100k} & \cellcolor{colorAlpacaTextLens_Inpt}{505} & \cellcolor{colorAlpacaTextLens_Tgt}{270} & \emojiblank\emoji{robot} & \greencheck & \emojiblank & \emojiblank & \emojiblank & \emojiblank & \TransparentCircle \TransparentCircle \NCDataCircle & \greencheck \\
Anthropic HH & \cellcolor{colorOpAsstOctoPackPropertyCounts_Datasets}{1} & \cellcolor{colorAnthropicHHPropertyCounts_Dialogs}{161k} & \cellcolor{colorOpAsstOctoPackPropertyCounts_Tasks}{3} & \cellcolor{colorTasksourceSTPropertyCounts_Langs}{1} & \cellcolor{colorOpAsstOctoPackPropertyCounts_Topics}{10} & \cellcolor{colorOpAsstOctoPackPropertyCounts_Domains}{1} & \cellcolor{colorAnthropicHHPropertyCounts_Downs}{82k} & \cellcolor{colorAnthropicHHTextLens_Inpt}{69} & \cellcolor{colorAnthropicHHTextLens_Tgt}{311} & \emojiblank\emoji{robot} & \emojiblank & \emojiblank & \emojiblank & \greencheck & \emojiblank & \CommercialDataCircle \TransparentCircle \TransparentCircle & \emojiblank \\
BaizeChat & \cellcolor{colorOpenOrcaPropertyCounts_Datasets}{4} & \cellcolor{colorBaizeChatPropertyCounts_Dialogs}{210k} & \cellcolor{colorCoTColl.PropertyCounts_Tasks}{12} & \cellcolor{colorStackExchangePropertyCounts_Langs}{2} & \cellcolor{colorBaizeChatPropertyCounts_Topics}{37} & \cellcolor{colorBaizeChatPropertyCounts_Domains}{3} & \cellcolor{colorBaizeChatPropertyCounts_Downs}{<1k} & \cellcolor{colorBaizeChatTextLens_Inpt}{74} & \cellcolor{colorBaizeChatTextLens_Tgt}{234} & \emojiblank\emoji{robot} & \greencheck & \emojiblank & \emojiblank & \emojiblank & \emojiblank & \TransparentCircle \TransparentCircle \NCDataCircle & \greencheck \\
BookSum & \cellcolor{colorOpAsstOctoPackPropertyCounts_Datasets}{1} & \cellcolor{colorBookSumPropertyCounts_Dialogs}{7k} & \cellcolor{colorTinyStoriesPropertyCounts_Tasks}{4} & \cellcolor{colorTasksourceSTPropertyCounts_Langs}{1} & \cellcolor{colorOpAsstOctoPackPropertyCounts_Topics}{10} & \cellcolor{colorOpAsstOctoPackPropertyCounts_Domains}{1} & \cellcolor{colorBookSumPropertyCounts_Downs}{<1k} & \cellcolor{colorBookSumTextLens_Inpt}{14k} & \cellcolor{colorBookSumTextLens_Tgt}{2k} & \emoji{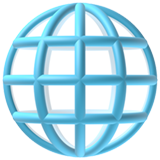}\emojiblank & \greencheck & \emojiblank & \emojiblank & \emojiblank & \emojiblank & \TransparentCircle \TransparentCircle \NCDataCircle & \emojiblank \\
CamelAI Sci. & \cellcolor{colorCamelAISci.PropertyCounts_Datasets}{3} & \cellcolor{colorCamelAISci.PropertyCounts_Dialogs}{60k} & \cellcolor{colorTool-LlamaPropertyCounts_Tasks}{2} & \cellcolor{colorTasksourceSTPropertyCounts_Langs}{1} & \cellcolor{colorCoTColl.PropertyCounts_Topics}{29} & \cellcolor{colorOpAsstOctoPackPropertyCounts_Domains}{1} & \cellcolor{colorCamelAISci.PropertyCounts_Downs}{<1k} & \cellcolor{colorCamelAISci.TextLens_Inpt}{190} & \cellcolor{colorCamelAISci.TextLens_Tgt}{2k} & \emojiblank\emoji{robot} & \greencheck & \emojiblank & \emojiblank & \emojiblank & \emojiblank & \TransparentCircle \TransparentCircle \NCDataCircle & \greencheck \\
CoT Coll. & \cellcolor{colorCoTColl.PropertyCounts_Datasets}{6} & \cellcolor{colorCoTColl.PropertyCounts_Dialogs}{2,183k} & \cellcolor{colorCoTColl.PropertyCounts_Tasks}{12} & \cellcolor{colorCoTColl.PropertyCounts_Langs}{7} & \cellcolor{colorCoTColl.PropertyCounts_Topics}{29} & \cellcolor{colorOpAsstOctoPackPropertyCounts_Domains}{1} & \cellcolor{colorCoTColl.PropertyCounts_Downs}{<1k} & \cellcolor{colorCoTColl.TextLens_Inpt}{728} & \cellcolor{colorCoTColl.TextLens_Tgt}{265} & \emojiblank\emoji{robot} & \emojiblank & \emojiblank & \greencheck & \emojiblank & \emojiblank & \TransparentCircle \TransparentCircle \NCDataCircle & \greencheck \\
Code Alpaca & \cellcolor{colorOpAsstOctoPackPropertyCounts_Datasets}{1} & \cellcolor{colorCodeAlpacaPropertyCounts_Dialogs}{20k} & \cellcolor{colorOpAsstOctoPackPropertyCounts_Tasks}{3} & \cellcolor{colorStackExchangePropertyCounts_Langs}{2} & \cellcolor{colorOpAsstOctoPackPropertyCounts_Topics}{10} & \cellcolor{colorOpAsstOctoPackPropertyCounts_Domains}{1} & \cellcolor{colorCodeAlpacaPropertyCounts_Downs}{5k} & \cellcolor{colorCodeAlpacaTextLens_Inpt}{97} & \cellcolor{colorCodeAlpacaTextLens_Tgt}{196} & \emojiblank\emoji{robot} & \greencheck & \emojiblank & \emojiblank & \emojiblank & \emojiblank & \TransparentCircle \UnspecifiedDataCircle \TransparentCircle & \greencheck \\
CommitPackFT & \cellcolor{colorCommitPackFTPropertyCounts_Datasets}{277} & \cellcolor{colorCommitPackFTPropertyCounts_Dialogs}{702k} & \cellcolor{colorCommitPackFTPropertyCounts_Tasks}{1} & \cellcolor{colorCommitPackFTPropertyCounts_Langs}{278} & \cellcolor{colorCommitPackFTPropertyCounts_Topics}{751} & \cellcolor{colorOpAsstOctoPackPropertyCounts_Domains}{1} & \cellcolor{colorCommitPackFTPropertyCounts_Downs}{4k} & \cellcolor{colorCommitPackFTTextLens_Inpt}{645} & \cellcolor{colorCommitPackFTTextLens_Tgt}{784} & \emoji{globe-with-meridians}\emojiblank & \greencheck & \emojiblank & \emojiblank & \emojiblank & \emojiblank & \CommercialDataCircle \UnspecifiedDataCircle \TransparentCircle & \emojiblank \\
Dolly 15k & \cellcolor{colorGPT4AllJPropertyCounts_Datasets}{7} & \cellcolor{colorDolly15kPropertyCounts_Dialogs}{15k} & \cellcolor{colorAiroborosPropertyCounts_Tasks}{5} & \cellcolor{colorTasksourceSTPropertyCounts_Langs}{1} & \cellcolor{colorDolly15kPropertyCounts_Topics}{38} & \cellcolor{colorOpAsstOctoPackPropertyCounts_Domains}{1} & \cellcolor{colorDolly15kPropertyCounts_Downs}{10,116k} & \cellcolor{colorDolly15kTextLens_Inpt}{423} & \cellcolor{colorDolly15kTextLens_Tgt}{357} & \emoji{globe-with-meridians}\emojiblank & \greencheck & \emojiblank & \emojiblank & \emojiblank & \emojiblank & \CommercialDataCircle \TransparentCircle \TransparentCircle & \emojiblank \\
Evol-Instr. & \cellcolor{colorEvol-Instr.PropertyCounts_Datasets}{2} & \cellcolor{colorEvol-Instr.PropertyCounts_Dialogs}{213k} & \cellcolor{colorEvol-Instr.PropertyCounts_Tasks}{11} & \cellcolor{colorStackExchangePropertyCounts_Langs}{2} & \cellcolor{colorEvol-Instr.PropertyCounts_Topics}{17} & \cellcolor{colorOpAsstOctoPackPropertyCounts_Domains}{1} & \cellcolor{colorEvol-Instr.PropertyCounts_Downs}{2k} & \cellcolor{colorEvol-Instr.TextLens_Inpt}{570} & \cellcolor{colorEvol-Instr.TextLens_Tgt}{2k} & \emojiblank\emoji{robot} & \greencheck & \emojiblank & \emojiblank & \emojiblank & \emojiblank & \TransparentCircle \TransparentCircle \NCDataCircle & \greencheck \\
Flan Collection & \cellcolor{colorFlanCollectionPropertyCounts_Datasets}{450} & \cellcolor{colorFlanCollectionPropertyCounts_Dialogs}{9,813k} & \cellcolor{colorFlanCollectionPropertyCounts_Tasks}{19} & \cellcolor{colorFlanCollectionPropertyCounts_Langs}{39} & \cellcolor{colorFlanCollectionPropertyCounts_Topics}{1k} & \cellcolor{colorOpenOrcaPropertyCounts_Domains}{23} & \cellcolor{colorFlanCollectionPropertyCounts_Downs}{19k} & \cellcolor{colorFlanCollectionTextLens_Inpt}{2k} & \cellcolor{colorFlanCollectionTextLens_Tgt}{128} & \emoji{globe-with-meridians}\emoji{robot} & \greencheck & \greencheck & \greencheck & \emojiblank & \emojiblank & \CommercialDataCircle \UnspecifiedDataCircle \NCDataCircle & \greencheck \\
GPT-4-Alpaca & \cellcolor{colorOpAsstOctoPackPropertyCounts_Datasets}{1} & \cellcolor{colorGPT-4-AlpacaPropertyCounts_Dialogs}{55k} & \cellcolor{colorUltraChatPropertyCounts_Tasks}{7} & \cellcolor{colorTasksourceSTPropertyCounts_Langs}{1} & \cellcolor{colorOpAsstOctoPackPropertyCounts_Topics}{10} & \cellcolor{colorOpAsstOctoPackPropertyCounts_Domains}{1} & \cellcolor{colorGPT-4-AlpacaPropertyCounts_Downs}{1k} & \cellcolor{colorGPT-4-AlpacaTextLens_Inpt}{130} & \cellcolor{colorGPT-4-AlpacaTextLens_Tgt}{543} & \emojiblank\emoji{robot} & \greencheck & \emojiblank & \emojiblank & \emojiblank & \emojiblank & \TransparentCircle \TransparentCircle \NCDataCircle & \greencheck \\
GPT4AllJ & \cellcolor{colorGPT4AllJPropertyCounts_Datasets}{7} & \cellcolor{colorGPT4AllJPropertyCounts_Dialogs}{809k} & \cellcolor{colorGPT4AllJPropertyCounts_Tasks}{10} & \cellcolor{colorTasksourceSTPropertyCounts_Langs}{1} & \cellcolor{colorGPT4AllJPropertyCounts_Topics}{56} & \cellcolor{colorOpAsstOctoPackPropertyCounts_Domains}{1} & \cellcolor{colorGPT4AllJPropertyCounts_Downs}{<1k} & \cellcolor{colorGPT4AllJTextLens_Inpt}{883} & \cellcolor{colorGPT4AllJTextLens_Tgt}{1k} & \emojiblank\emoji{robot} & \greencheck & \emojiblank & \emojiblank & \emojiblank & \emojiblank & \TransparentCircle \UnspecifiedDataCircle \NCDataCircle & \greencheck \\
GPTeacher & \cellcolor{colorOpenOrcaPropertyCounts_Datasets}{4} & \cellcolor{colorGPTeacherPropertyCounts_Dialogs}{103k} & \cellcolor{colorGPTeacherPropertyCounts_Tasks}{8} & \cellcolor{colorStackExchangePropertyCounts_Langs}{2} & \cellcolor{colorGPTeacherPropertyCounts_Topics}{33} & \cellcolor{colorOpAsstOctoPackPropertyCounts_Domains}{1} & \cellcolor{colorGPTeacherPropertyCounts_Downs}{<1k} & \cellcolor{colorGPTeacherTextLens_Inpt}{227} & \cellcolor{colorGPTeacherTextLens_Tgt}{360} & \emojiblank\emoji{robot} & \greencheck & \emojiblank & \emojiblank & \emojiblank & \emojiblank & \CommercialDataCircle \TransparentCircle \TransparentCircle & \greencheck \\
Gorilla & \cellcolor{colorOpAsstOctoPackPropertyCounts_Datasets}{1} & \cellcolor{colorGorillaPropertyCounts_Dialogs}{15k} & \cellcolor{colorTinyStoriesPropertyCounts_Tasks}{4} & \cellcolor{colorStackExchangePropertyCounts_Langs}{2} & \cellcolor{colorOpAsstOctoPackPropertyCounts_Topics}{10} & \cellcolor{colorUltraChatPropertyCounts_Domains}{2} & \cellcolor{colorGorillaPropertyCounts_Downs}{<1k} & \cellcolor{colorHC3TextLens_Inpt}{119} & \cellcolor{colorGorillaTextLens_Tgt}{76} & \emojiblank\emoji{robot} & \greencheck & \emojiblank & \emojiblank & \emojiblank & \emojiblank & \CommercialDataCircle \TransparentCircle \TransparentCircle & \greencheck \\
HC3 & \cellcolor{colorHC3PropertyCounts_Datasets}{12} & \cellcolor{colorHC3PropertyCounts_Dialogs}{37k} & \cellcolor{colorHC3PropertyCounts_Tasks}{6} & \cellcolor{colorStackExchangePropertyCounts_Langs}{2} & \cellcolor{colorHC3PropertyCounts_Topics}{102} & \cellcolor{colorLIMAPropertyCounts_Domains}{6} & \cellcolor{colorHC3PropertyCounts_Downs}{2k} & \cellcolor{colorHC3TextLens_Inpt}{119} & \cellcolor{colorHC3TextLens_Tgt}{652} & \emojiblank\emoji{robot} & \emojiblank & \emojiblank & \emojiblank & \greencheck & \emojiblank & \CommercialDataCircle \UnspecifiedDataCircle \NCDataCircle & \greencheck \\
Joke Expl. & \cellcolor{colorOpAsstOctoPackPropertyCounts_Datasets}{1} & \cellcolor{colorJokeExpl.PropertyCounts_Dialogs}{<1k} & \cellcolor{colorTool-LlamaPropertyCounts_Tasks}{2} & \cellcolor{colorTasksourceSTPropertyCounts_Langs}{1} & \cellcolor{colorOpAsstOctoPackPropertyCounts_Topics}{10} & \cellcolor{colorOpAsstOctoPackPropertyCounts_Domains}{1} & \cellcolor{colorJokeExpl.PropertyCounts_Downs}{<1k} & \cellcolor{colorJokeExpl.TextLens_Inpt}{96} & \cellcolor{colorJokeExpl.TextLens_Tgt}{547} & \emoji{globe-with-meridians}\emojiblank & \greencheck & \emojiblank & \emojiblank & \emojiblank & \emojiblank & \CommercialDataCircle \TransparentCircle \TransparentCircle & \emojiblank \\
LAION OIG & \cellcolor{colorLAIONOIGPropertyCounts_Datasets}{26} & \cellcolor{colorLAIONOIGPropertyCounts_Dialogs}{9,211k} & \cellcolor{colorCoTColl.PropertyCounts_Tasks}{12} & \cellcolor{colorTasksourceSTPropertyCounts_Langs}{1} & \cellcolor{colorLAIONOIGPropertyCounts_Topics}{171} & \cellcolor{colorLAIONOIGPropertyCounts_Domains}{11} & \cellcolor{colorLAIONOIGPropertyCounts_Downs}{<1k} & \cellcolor{colorLAIONOIGTextLens_Inpt}{343} & \cellcolor{colorLAIONOIGTextLens_Tgt}{595} & \emoji{globe-with-meridians}\emoji{robot} & \emojiblank & \emojiblank & \emojiblank & \emojiblank & \greencheck & \CommercialDataCircle \UnspecifiedDataCircle \TransparentCircle & \greencheck \\
LIMA & \cellcolor{colorLIMAPropertyCounts_Datasets}{5} & \cellcolor{colorLIMAPropertyCounts_Dialogs}{1k} & \cellcolor{colorGPT4AllJPropertyCounts_Tasks}{10} & \cellcolor{colorStackExchangePropertyCounts_Langs}{2} & \cellcolor{colorLIMAPropertyCounts_Topics}{43} & \cellcolor{colorLIMAPropertyCounts_Domains}{6} & \cellcolor{colorLIMAPropertyCounts_Downs}{3k} & \cellcolor{colorLIMATextLens_Inpt}{228} & \cellcolor{colorLIMATextLens_Tgt}{3k} & \emoji{globe-with-meridians}\emojiblank & \greencheck & \greencheck & \emojiblank & \emojiblank & \greencheck & \TransparentCircle \TransparentCircle \NCDataCircle & \emojiblank \\
Longform & \cellcolor{colorGPT4AllJPropertyCounts_Datasets}{7} & \cellcolor{colorLongformPropertyCounts_Dialogs}{23k} & \cellcolor{colorEvol-Instr.PropertyCounts_Tasks}{11} & \cellcolor{colorTasksourceSTPropertyCounts_Langs}{1} & \cellcolor{colorLongformPropertyCounts_Topics}{63} & \cellcolor{colorLongformPropertyCounts_Domains}{4} & \cellcolor{colorLongformPropertyCounts_Downs}{3k} & \cellcolor{colorLongformTextLens_Inpt}{810} & \cellcolor{colorLongformTextLens_Tgt}{2k} & \emojiblank\emoji{robot} & \greencheck & \emojiblank & \emojiblank & \emojiblank & \emojiblank & \CommercialDataCircle \UnspecifiedDataCircle \TransparentCircle & \greencheck \\
OpAsst OctoPack & \cellcolor{colorOpAsstOctoPackPropertyCounts_Datasets}{1} & \cellcolor{colorOpAsstOctoPackPropertyCounts_Dialogs}{10k} & \cellcolor{colorOpAsstOctoPackPropertyCounts_Tasks}{3} & \cellcolor{colorOpAsstOctoPackPropertyCounts_Langs}{20} & \cellcolor{colorOpAsstOctoPackPropertyCounts_Topics}{10} & \cellcolor{colorOpAsstOctoPackPropertyCounts_Domains}{1} & \cellcolor{colorOpAsstOctoPackPropertyCounts_Downs}{<1k} & \cellcolor{colorOpAsstOctoPackTextLens_Inpt}{118} & \cellcolor{colorOpAsstOctoPackTextLens_Tgt}{884} & \emoji{globe-with-meridians}\emojiblank & \emojiblank & \emojiblank & \emojiblank & \emojiblank & \greencheck & \CommercialDataCircle \TransparentCircle \TransparentCircle & \emojiblank \\
OpenAI Summ. & \cellcolor{colorOpAsstOctoPackPropertyCounts_Datasets}{1} & \cellcolor{colorOpenAISumm.PropertyCounts_Dialogs}{93k} & \cellcolor{colorAiroborosPropertyCounts_Tasks}{5} & \cellcolor{colorTasksourceSTPropertyCounts_Langs}{1} & \cellcolor{colorOpAsstOctoPackPropertyCounts_Topics}{10} & \cellcolor{colorOpAsstOctoPackPropertyCounts_Domains}{1} & \cellcolor{colorOpenAISumm.PropertyCounts_Downs}{14k} & \cellcolor{colorOpenAISumm.TextLens_Inpt}{1k} & \cellcolor{colorOpenAISumm.TextLens_Tgt}{134} & \emojiblank\emoji{robot} & \emojiblank & \emojiblank & \emojiblank & \greencheck & \emojiblank & \CommercialDataCircle \TransparentCircle \TransparentCircle & \greencheck \\
OpenAssistant & \cellcolor{colorOpenAssistantPropertyCounts_Datasets}{19} & \cellcolor{colorOpenAssistantPropertyCounts_Dialogs}{10k} & \cellcolor{colorTinyStoriesPropertyCounts_Tasks}{4} & \cellcolor{colorOpAsstOctoPackPropertyCounts_Langs}{20} & \cellcolor{colorOpenAssistantPropertyCounts_Topics}{99} & \cellcolor{colorOpAsstOctoPackPropertyCounts_Domains}{1} & \cellcolor{colorOpenAssistantPropertyCounts_Downs}{14k} & \cellcolor{colorOpAsstOctoPackTextLens_Inpt}{118} & \cellcolor{colorOpenAssistantTextLens_Tgt}{711} & \emoji{globe-with-meridians}\emojiblank & \emojiblank & \emojiblank & \emojiblank & \emojiblank & \greencheck & \CommercialDataCircle \TransparentCircle \TransparentCircle & \emojiblank \\
OpenOrca & \cellcolor{colorOpenOrcaPropertyCounts_Datasets}{4} & \cellcolor{colorOpenOrcaPropertyCounts_Dialogs}{4,234k} & \cellcolor{colorEvol-Instr.PropertyCounts_Tasks}{11} & \cellcolor{colorTasksourceSTPropertyCounts_Langs}{1} & \cellcolor{colorOpenOrcaPropertyCounts_Topics}{30} & \cellcolor{colorOpenOrcaPropertyCounts_Domains}{23} & \cellcolor{colorOpenOrcaPropertyCounts_Downs}{28k} & \cellcolor{colorOpenOrcaTextLens_Inpt}{1k} & \cellcolor{colorOpenOrcaTextLens_Tgt}{492} & \emojiblank\emoji{robot} & \greencheck & \emojiblank & \emojiblank & \emojiblank & \emojiblank & \CommercialDataCircle \TransparentCircle \NCDataCircle & \greencheck \\
SHP & \cellcolor{colorSHPPropertyCounts_Datasets}{18} & \cellcolor{colorSHPPropertyCounts_Dialogs}{349k} & \cellcolor{colorHC3PropertyCounts_Tasks}{6} & \cellcolor{colorStackExchangePropertyCounts_Langs}{2} & \cellcolor{colorSHPPropertyCounts_Topics}{151} & \cellcolor{colorOpAsstOctoPackPropertyCounts_Domains}{1} & \cellcolor{colorSHPPropertyCounts_Downs}{4k} & \cellcolor{colorSHPTextLens_Inpt}{824} & \cellcolor{colorSHPTextLens_Tgt}{496} & \emoji{globe-with-meridians}\emojiblank & \emojiblank & \emojiblank & \emojiblank & \greencheck & \emojiblank & \TransparentCircle \UnspecifiedDataCircle \TransparentCircle & \emojiblank \\
Self-Instruct & \cellcolor{colorOpAsstOctoPackPropertyCounts_Datasets}{1} & \cellcolor{colorSelf-InstructPropertyCounts_Dialogs}{83k} & \cellcolor{colorHC3PropertyCounts_Tasks}{6} & \cellcolor{colorStackExchangePropertyCounts_Langs}{2} & \cellcolor{colorOpAsstOctoPackPropertyCounts_Topics}{10} & \cellcolor{colorOpAsstOctoPackPropertyCounts_Domains}{1} & \cellcolor{colorSelf-InstructPropertyCounts_Downs}{3k} & \cellcolor{colorSelf-InstructTextLens_Inpt}{134} & \cellcolor{colorSelf-InstructTextLens_Tgt}{104} & \emojiblank\emoji{robot} & \greencheck & \emojiblank & \emojiblank & \emojiblank & \emojiblank & \CommercialDataCircle \TransparentCircle \TransparentCircle & \greencheck \\
ShareGPT & \cellcolor{colorOpAsstOctoPackPropertyCounts_Datasets}{1} & \cellcolor{colorShareGPTPropertyCounts_Dialogs}{77k} & \cellcolor{colorShareGPTPropertyCounts_Tasks}{9} & \cellcolor{colorTasksourceSTPropertyCounts_Langs}{1} & \cellcolor{colorOpAsstOctoPackPropertyCounts_Topics}{10} & \cellcolor{colorUltraChatPropertyCounts_Domains}{2} & \cellcolor{colorShareGPTPropertyCounts_Downs}{<1k} & \cellcolor{colorShareGPTTextLens_Inpt}{303} & \cellcolor{colorShareGPTTextLens_Tgt}{1k} & \emojiblank\emoji{robot} & \emojiblank & \emojiblank & \emojiblank & \emojiblank & \greencheck & \TransparentCircle \UnspecifiedDataCircle \TransparentCircle & \greencheck \\
StackExchange & \cellcolor{colorOpAsstOctoPackPropertyCounts_Datasets}{1} & \cellcolor{colorStackExchangePropertyCounts_Dialogs}{10,607k} & \cellcolor{colorCommitPackFTPropertyCounts_Tasks}{1} & \cellcolor{colorStackExchangePropertyCounts_Langs}{2} & \cellcolor{colorOpAsstOctoPackPropertyCounts_Topics}{10} & \cellcolor{colorOpAsstOctoPackPropertyCounts_Domains}{1} & \cellcolor{colorStackExchangePropertyCounts_Downs}{<1k} & \cellcolor{colorStackExchangeTextLens_Inpt}{1k} & \cellcolor{colorStackExchangeTextLens_Tgt}{901} & \emoji{globe-with-meridians}\emojiblank & \greencheck & \emojiblank & \emojiblank & \emojiblank & \emojiblank & \TransparentCircle \UnspecifiedDataCircle \TransparentCircle & \emojiblank \\
StarCoder & \cellcolor{colorOpAsstOctoPackPropertyCounts_Datasets}{1} & \cellcolor{colorStarCoderPropertyCounts_Dialogs}{<1k} & \cellcolor{colorCommitPackFTPropertyCounts_Tasks}{1} & \cellcolor{colorStackExchangePropertyCounts_Langs}{2} & \cellcolor{colorOpAsstOctoPackPropertyCounts_Topics}{10} & \cellcolor{colorOpAsstOctoPackPropertyCounts_Domains}{1} & \cellcolor{colorStarCoderPropertyCounts_Downs}{<1k} & \cellcolor{colorStarCoderTextLens_Inpt}{195} & \cellcolor{colorStarCoderTextLens_Tgt}{504} & \emojiblank\emoji{robot} & \greencheck & \emojiblank & \emojiblank & \emojiblank & \emojiblank & \CommercialDataCircle \TransparentCircle \TransparentCircle & \emojiblank \\
Tasksource Ins. & \cellcolor{colorTasksourceIns.PropertyCounts_Datasets}{288} & \cellcolor{colorTasksourceIns.PropertyCounts_Dialogs}{3,397k} & \cellcolor{colorTasksourceIns.PropertyCounts_Tasks}{13} & \cellcolor{colorTasksourceSTPropertyCounts_Langs}{1} & \cellcolor{colorTasksourceIns.PropertyCounts_Topics}{582} & \cellcolor{colorTasksourceIns.PropertyCounts_Domains}{20} & \cellcolor{colorTasksourceIns.PropertyCounts_Downs}{<1k} & \cellcolor{colorTasksourceIns.TextLens_Inpt}{518} & \cellcolor{colorTasksourceIns.TextLens_Tgt}{18} & \emoji{globe-with-meridians}\emoji{robot} & \greencheck & \emojiblank & \emojiblank & \emojiblank & \emojiblank & \CommercialDataCircle \UnspecifiedDataCircle \NCDataCircle & \greencheck \\
Tasksource ST & \cellcolor{colorTasksourceSTPropertyCounts_Datasets}{229} & \cellcolor{colorTasksourceSTPropertyCounts_Dialogs}{338k} & \cellcolor{colorTasksourceSTPropertyCounts_Tasks}{15} & \cellcolor{colorTasksourceSTPropertyCounts_Langs}{1} & \cellcolor{colorTasksourceSTPropertyCounts_Topics}{477} & \cellcolor{colorTasksourceSTPropertyCounts_Domains}{18} & \cellcolor{colorTasksourceSTPropertyCounts_Downs}{<1k} & \cellcolor{colorTasksourceSTTextLens_Inpt}{3k} & \cellcolor{colorTasksourceSTTextLens_Tgt}{6} & \emoji{globe-with-meridians}\emoji{robot} & \greencheck & \emojiblank & \emojiblank & \emojiblank & \emojiblank & \CommercialDataCircle \UnspecifiedDataCircle \NCDataCircle & \greencheck \\
TinyStories & \cellcolor{colorOpAsstOctoPackPropertyCounts_Datasets}{1} & \cellcolor{colorTinyStoriesPropertyCounts_Dialogs}{14k} & \cellcolor{colorTinyStoriesPropertyCounts_Tasks}{4} & \cellcolor{colorTasksourceSTPropertyCounts_Langs}{1} & \cellcolor{colorOpAsstOctoPackPropertyCounts_Topics}{10} & \cellcolor{colorOpAsstOctoPackPropertyCounts_Domains}{1} & \cellcolor{colorTinyStoriesPropertyCounts_Downs}{12k} & \cellcolor{colorTinyStoriesTextLens_Inpt}{517} & \cellcolor{colorTinyStoriesTextLens_Tgt}{194k} & \emojiblank\emoji{robot} & \greencheck & \emojiblank & \emojiblank & \emojiblank & \emojiblank & \CommercialDataCircle \TransparentCircle \TransparentCircle & \greencheck \\
Tool-Llama & \cellcolor{colorOpAsstOctoPackPropertyCounts_Datasets}{1} & \cellcolor{colorTool-LlamaPropertyCounts_Dialogs}{37k} & \cellcolor{colorTool-LlamaPropertyCounts_Tasks}{2} & \cellcolor{colorStackExchangePropertyCounts_Langs}{2} & \cellcolor{colorOpAsstOctoPackPropertyCounts_Topics}{10} & \cellcolor{colorOpAsstOctoPackPropertyCounts_Domains}{1} & - & \cellcolor{colorTool-LlamaTextLens_Inpt}{7k} & \cellcolor{colorTool-LlamaTextLens_Tgt}{1k} & \emojiblank\emoji{robot} & \emojiblank & \emojiblank & \emojiblank & \emojiblank & \greencheck & \TransparentCircle \TransparentCircle \NCDataCircle & \greencheck \\
UltraChat & \cellcolor{colorOpAsstOctoPackPropertyCounts_Datasets}{1} & \cellcolor{colorUltraChatPropertyCounts_Dialogs}{1,468k} & \cellcolor{colorUltraChatPropertyCounts_Tasks}{7} & \cellcolor{colorTasksourceSTPropertyCounts_Langs}{1} & \cellcolor{colorUltraChatPropertyCounts_Topics}{11} & \cellcolor{colorUltraChatPropertyCounts_Domains}{2} & \cellcolor{colorUltraChatPropertyCounts_Downs}{2k} & \cellcolor{colorUltraChatTextLens_Inpt}{282} & \cellcolor{colorUltraChatTextLens_Tgt}{1k} & \emojiblank\emoji{robot} & \greencheck & \emojiblank & \emojiblank & \emojiblank & \greencheck & \TransparentCircle \TransparentCircle \NCDataCircle & \greencheck \\
Unnatural Instr. & \cellcolor{colorOpAsstOctoPackPropertyCounts_Datasets}{1} & \cellcolor{colorUnnaturalInstr.PropertyCounts_Dialogs}{66k} & \cellcolor{colorTinyStoriesPropertyCounts_Tasks}{4} & \cellcolor{colorTasksourceSTPropertyCounts_Langs}{1} & \cellcolor{colorOpAsstOctoPackPropertyCounts_Topics}{10} & \cellcolor{colorOpAsstOctoPackPropertyCounts_Domains}{1} & \cellcolor{colorUnnaturalInstr.PropertyCounts_Downs}{<1k} & \cellcolor{colorUnnaturalInstr.TextLens_Inpt}{331} & \cellcolor{colorUnnaturalInstr.TextLens_Tgt}{68} & \emojiblank\emoji{robot} & \greencheck & \emojiblank & \emojiblank & \emojiblank & \emojiblank & \CommercialDataCircle \TransparentCircle \TransparentCircle & \greencheck \\
WebGPT & \cellcolor{colorLIMAPropertyCounts_Datasets}{5} & \cellcolor{colorWebGPTPropertyCounts_Dialogs}{20k} & \cellcolor{colorTinyStoriesPropertyCounts_Tasks}{4} & \cellcolor{colorTasksourceSTPropertyCounts_Langs}{1} & \cellcolor{colorWebGPTPropertyCounts_Topics}{35} & \cellcolor{colorBaizeChatPropertyCounts_Domains}{3} & \cellcolor{colorWebGPTPropertyCounts_Downs}{1k} & \cellcolor{colorWebGPTTextLens_Inpt}{737} & \cellcolor{colorWebGPTTextLens_Tgt}{743} & \emojiblank\emoji{robot} & \emojiblank & \emojiblank & \emojiblank & \greencheck & \emojiblank & \CommercialDataCircle \TransparentCircle \TransparentCircle & \greencheck \\
xP3x & \cellcolor{colorxP3xPropertyCounts_Datasets}{467} & \cellcolor{colorxP3xPropertyCounts_Dialogs}{886,240k} & \cellcolor{colorAiroborosPropertyCounts_Tasks}{5} & \cellcolor{colorxP3xPropertyCounts_Langs}{245} & \cellcolor{colorSHPPropertyCounts_Topics}{151} & \cellcolor{colorxP3xPropertyCounts_Domains}{14} & \cellcolor{colorxP3xPropertyCounts_Downs}{<1k} & \cellcolor{colorxP3xTextLens_Inpt}{589} & \cellcolor{colorxP3xTextLens_Tgt}{441} & \emoji{globe-with-meridians}\emoji{robot} & \greencheck & \emojiblank & \emojiblank & \emojiblank & \emojiblank & \CommercialDataCircle \UnspecifiedDataCircle \NCDataCircle & \emojiblank \\
\bottomrule
\end{tabular}
}
\caption{\textbf{Alignment tuning collections and their characteristics.}
Properties of the collections include the numbers of datasets, dialogs, unique tasks, languages, topics, text domains, Huggingface monthly downloads (“Downs”), and the average length of input and target text, by characters. The \textsc{Source} column indicates whether a collection includes human web text (\emoji{globe-with-meridians}), or model generated text (\emoji{robot}). The dialog formats of each collection can be: zero-shot (Z), few-shot (F), chain-of-thought (C), response ranking (R), and multi-turn
dialog (M). The \textsc{Use} column indicates whether a collection includes data licensed for commercial use (\protect\CommercialDataCircle), data with no license (“unspecified”: \protect\UnspecifiedDataCircle), data only licensed for non-commercial or academic use (\protect\NCDataCircle). \emph{Note that these licenses are self-reported and their applicability is complicated, requiring legal consultation.}
The ``O'' column indicates if the collection includes OpenAI model generations, which may or may not affect commercial viability (see \cref{llm-generation})}
\label{tab:collections}
\end{table*}

\vspace{-1mm}
\subsection{Data Provenance Explorer (\dpiex)}
\label{sec:data-attributes}
Our information audit spans (I) \emph{identifier information}, bridging metadata from several aggregators, including Hugging Face, GitHub, Papers with Code, Semantic Scholar, and ArXiv, (II) detailed \emph{dataset characteristics} for a richer understanding of training set composition, and (III) \emph{dataset provenance} for licensing and attribution.
We expand our provenance metadata beyond just licenses, because conversations with practitioners revealed they rely not only on data licenses, but on a specific \textit{legal \& ethical risk tolerance}, parameterized by (a) the lineage of licenses, (b) the data source, (c) the creator's identity, and (d) the precedence of adoption by other developers. 

We release our extensive audit, as two tools: (1) a data explorer interface, the \emph{Data Provenance Explorer (\dpiex)} for widespread use, and (2) an accompanying repository for practitioners to download the data filtered for license conditions.
Practitioners are also able to generate a human-readable, markdown summary, or \emph{Data Provenance Card}, of the used datasets, and compositional properties for languages, tasks, and licenses (\cref{sec:data-prov-card}).
Modern researchers training on hundreds of datasets often find it onerous to manually curate extensive data cards for these compilations\citep{mitchell2019model,gebru2021datasheets}. 
We hope this tool will aid in writing the data attribution and composition sections of these documentation efforts, by providing auto-generated, copy-and-pastable dataframe summaries.

Collecting comprehensive metadata for each dataset required leveraging several sources including collection by linking to resources already on the web (\emoji{globe-with-meridians}), human annotation by legal experts (\emoji{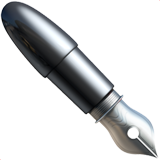}), or using GPT-4 to assist in human annotation (\emoji{robot}). 


\vspace{-2mm}
\paragraph{Identifier Information} discloses links and connects aggregator identifiers.
\vspace{-1mm}
\begin{enumerate}\itemsep0em
  \item \textbf{Dataset Identifiers \emoji{black-nib}}: The dataset's name, associated paper title, and description of the dataset.
  \item \textbf{Dataset Aggregator Links \emoji{black-nib}}: A link to each major aggregator, including GitHub, Hugging Face, Papers with Code, Semantic Scholar, and ArXiv allows us to incorporate and compare their crowdsourced metadata.
  \item \textbf{Collection \emoji{black-nib} }: The name and URL to the data collection of which this dataset is a part.
\end{enumerate} 

\vspace{-2mm}
\paragraph{Dataset Characteristics} detail information relevant to understanding data representation/composition, and curating a training set.
\vspace{-1mm}
\begin{enumerate}\itemsep0em
  \item \textbf{Languages \emoji{black-nib}}: Each of the languages represented in the dataset, so developers can easily follow the ``Bender Rule'' \citep{bender2011achieving}.
  \item \textbf{Task Categories \emoji{black-nib}\emoji{robot}}: The 20+ task categories represented in the instructions, such as Question Answering, Translation, Program Synthesis, Toxicity Identification, Creative Writing, and Roleplaying.
  \item \textbf{Text Topics \emoji{robot}}: An automated annotation of the topics discussed in the datasets, with GPT-4 labeling a sample of 100 examples for up to 10 covered topics. 
  \item \textbf{Text Length Metrics }: The minimum, maximum, and mean number of dialog turns per conversation, of characters (agnostic to tokenization/non-whitespace languages, as this introduces biases \citep{petrov2023language}) per user instruction and assistant responses.
  \item \textbf{Format \emoji{black-nib} }: The format and intended use of the data. The options are zero-shot prompts, few-shot prompts, chain-of-thought prompts, multi-turn dialog, and response ranking.
  \item \textbf{Time of Collection \emoji{globe-with-meridians} }: The time as which the work was published, which acts as an upper bound estimate of the age of the text. 
\end{enumerate} 

\vspace{-2mm}
\paragraph{Dataset Provenance}
\vspace{-1mm}
\begin{enumerate}\itemsep0em
  \item \textbf{Licenses \emoji{globe-with-meridians}\emoji{black-nib}}: The license name and URLs associated with the data, using the process described in \cref{sec:license-collection}. We also enable filtering by license use classes, categorized by legal professionals.
  \item \textbf{Text Source \emoji{black-nib}\emoji{robot}}: The original sources of the text, often Wikipedia, Reddit, or other scraped online/offline sources.
  \item \textbf{Creators \emoji{black-nib}}: The institutions of the dataset authors, including universities, corporations, and other organizations.
  \item \textbf{Attribution \emoji{globe-with-meridians}}: The attribution information for the authors of the paper associated with the dataset.
  \item \textbf{Citation \& Download Counts \emoji{globe-with-meridians}}: The citation and Hugging Face download count for the paper and dataset, dated September 2023. This acts as an estimate of community use, and is commonly used as precedence to decide on the risk level for using these datasets .
\end{enumerate} 

\subsection{License Annotation Process}
\label{sec:license-collection}

One of our central contributions is to validate the licenses associated with widely used and adopted datasets. 
This followed a time-intensive human annotation protocol, to collect dataset authors' self-reported licenses, and categorize them according to stated conditions.
Note that this protocol reflects best efforts to verify self-reported licenses, and does not constitute legal advice (see~\cref{sec:not-legal-advice}).
Additionally, it is important to note that the enforceability of these licenses depends on several factors discussed in~\cref{sec:legal-discussion}.
One especially important assumption in cases where datasets are based on data obtained from other sources is that dataset creators actually have a copyright interest in their dataset.
This depends on the data source and how creators modify or augment this data, and requires a case-by-case analysis.
However, it appears that most developers operate under the general assumption that they alone own their datasets.
Our license annotation workflow follows these steps:
\begin{enumerate}\itemsep0em
  \item \textbf{Compile all Self-Reported License Information} We aggregate all licensing information reported on GitHub, ArXiv, Hugging Face, Papers with Code, and the collection itself (e.g. Super-Natural Instructions, \citet{wang2022super}).
  \item \textbf{Search for explicit Data Licenses} The annotator searches for a license specifically given to the dataset (\emph{not the accompanying code}) by the authors.
  A license is found if (a) the GitHub repository mentions or links a license in reference to the data, (b) the Hugging Face license label was uploaded by the dataset creator themselves, (c) the paper, Hugging Face, or Papers with Code provide a dataset-specific license link, attributable to the data authors. 
  \item \textbf{Identify a License Type} A license may fall into a set of common types (e.g. MIT, Apache 2, CC BY SA, etc.), be a ``Custom'' license, a permission Request Form, or if none was found for the data, \emph{Unspecified}.
  If a dataset has multiple licenses, the annotator will list each of them, according to their types.
  \item \textbf{Categorize Licenses} 
  From the perspective of a machine learning practitioner, licensing typically is viewed through the lens of how it impacts the model lifecycle---does it impede or allow for training on the data, downstream use conditions, attributing, modifying or re-distributing it.
  Based on discussions with industry experts, we categorize licenses based on three important features that impact the model lifecycle: is data usage limited to academic or non-commercial purposes (\textbf{Permitted Use}), does the data source need to be attributed (\textbf{Attribution}), and do derivatives of the data need to licensed under the same terms as the original (\textbf{Share-Alike}).
  If there are multiple licenses for a dataset, its categorization for each feature is the chosen as the strictest across licenses.
  \item \textbf{Additional Provenance} In practice, legal teams may wish to balance their risk tolerance with more nuanced criteria. 
  For instance, they may be satisfied with using (more permissive) GitHub licenses, even when it is ambiguous whether these apply to the code or the data. They may also wish to include or exclude datasets based on whether these are already widely used in practice, where the original data was sourced from, and if the creator is a competitor. 
  To supplement the above license categories, we also collect all this metadata for fine-grained selection and filtering.
\end{enumerate}
\vspace{-1mm}

\begin{figure}[ht]
\centering
     \includegraphics[width=0.99\textwidth]{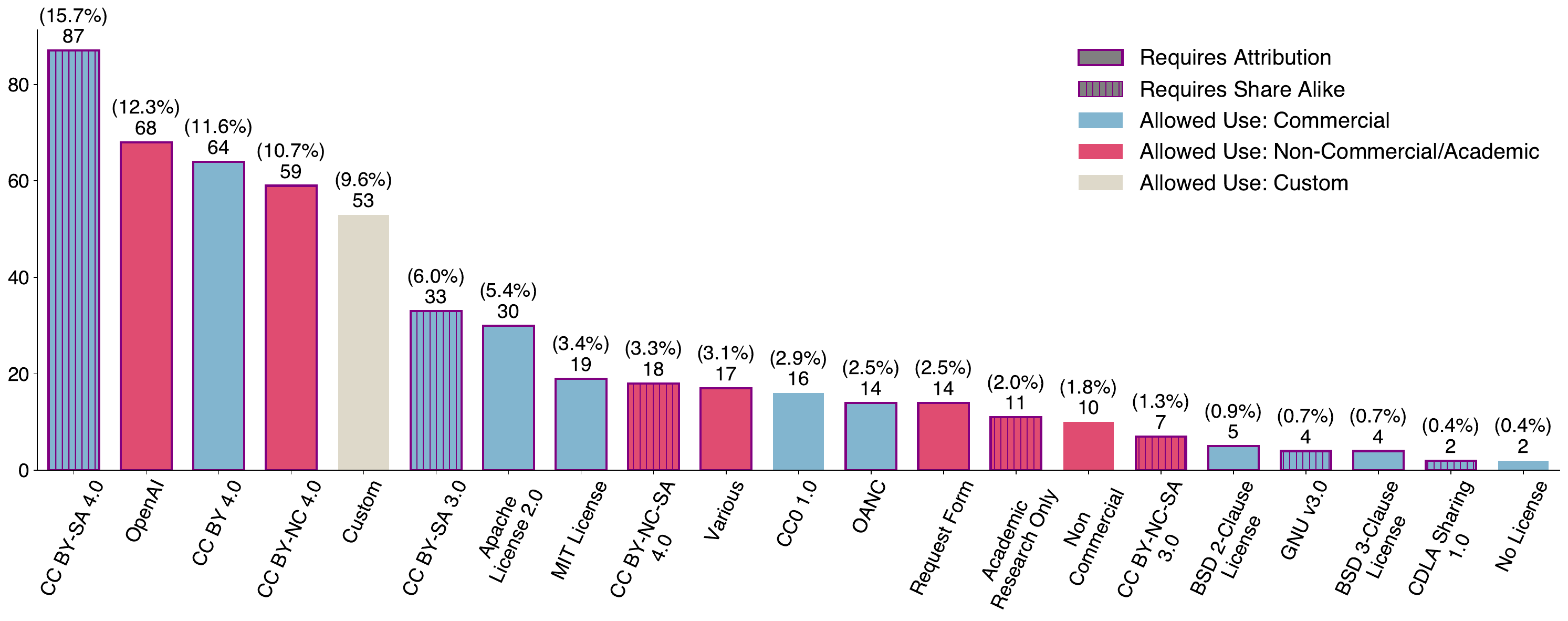}
      \caption{\textbf{We plot the distributions of licenses used in the~\dpic{}, a popular sample of the major supervised NLP datasets.} 
      We find a long tail of custom licenses, adopted from software for data.
      73\% of all licenses require attribution, and 33\% share-alike, but the most popular are usually commercially permissive.}
\label{fig:license-dist}
\vspace{-3mm}
\end{figure}

\vspace{-2mm}
\subsection{Data Provenance Card---\emph{A Data Bibliography}}
\label{sec:data-prov-card}

Prior work has stressed the importance of data documentation and attribution~\citep{bender-friedman-2018-data, bommasani2023foundation}.
In particular,~\citet{gebru2021datasheets}'s Datasheets breaks down documentation into motivation, composition, collection process, processing, uses, maintanence, and distribution. Similarly,~\citet{bender-friedman-2018-data} ask for curation rationale, language variety, speaker demographic, annotator demographic, speech situation, and text characteristics, among others.
However, when models train on many sources of data, even if they are each rigorously documented for each of these fields (rarely the case), it is challenging to cleanly synthesize comprehensive and navigable documentation for the resulting bundle.

To make this process tractable with scale, we propose leveraging \textbf{\emph{Symbolic Attribution}}, where our tools auto-generate a structured store of the provenance and attribution metadata, similar to a bibliography for data.\footnote{Auto-generated at \url{https://github.com/Data-Provenance-Initiative/Data-Provenance-Collection}}
Our collected schema allows this store to succinctly capture the attribution (links to repositories, aggregator copies, papers, creators), provenance (text/machine sources, licenses), and compositional properties of the data (languages, tasks, text metrics, format, and time).
This file of references and metadata, known as a \emph{Data Provenance Card} enables comprehensive documentation, proposed by prior work, while providing some advantages from its structure.
First, the Data Provenance Card can be easily searched, sorted, filtered and analyzed, whereas Datasheets or Statements, designed for individual datasets, are meant to be manually read.
Second, developers can efficiently assemble relevant information without losing any detail, by symbolically linking to the original datasets and their documentation.
Third, as datasets are continually re-packaged and absorbed into newer and bigger collections, Data Provenance Cards are easily adaptable by simply appending or concatenating them together.
Altogether, we hope this tooling enables and promotes the thorough documentation proposed in prior work~\citep{bender-friedman-2018-data,gebru2021datasheets,mitchell2019model, pushkarna2022data}

\vspace{-2mm}
\section{Empirical Analysis of Data Provenance}

\definecolor{colorCorrect}{RGB}{227,244,221}
\definecolor{colorIncorrect}{RGB}{247, 220, 222}

\begin{table*}
    \centering
    \begin{tabular}{lc||c|cccc}
        \toprule
        \multicolumn{2}{c||}{\textsc{\textbf{Correct license}}} & \multicolumn{5}{c}{\textsc{\textbf{License according to aggregators (Agg.)}}} \\
        \textsc{License} & \textsc{Count} & \textsc{Agg.} & \textsc{~~~~Comm.~~~~} & \textsc{~~~Unspec.~~~} & \textsc{Non-Comm.} & \textsc{Acad.-Only} \\
        \midrule

        \multirow{3}{*}{Commercial} & \multirow{3}{*}{\shortstack[l]{~~~856\\\small{(46.1\%)}}} & \github & \cellcolor{colorCorrect}{349} & 507 & 0 & 0 \\
        & & \hf & \cellcolor{colorCorrect}{176} & 677 & 1 & 2 \\
        & & \pwc & \cellcolor{colorCorrect}{313} & 520 & 1 & 22 \\
        \midrule        
        \multirow{3}{*}{Unspecified} & \multirow{3}{*}{\shortstack[l]{~~~570\\\small{(30.7\%)}}} & \github & \cellcolor{colorIncorrect}{112} & \cellcolor{colorCorrect}{458} & 0 & 0 \\
        & & \hf & \cellcolor{colorIncorrect}{164} & \cellcolor{colorCorrect}{395} & 6 & 5 \\
        & & \pwc & \cellcolor{colorIncorrect}{31} & \cellcolor{colorCorrect}{523} & 1 & 15 \\
        \midrule
        \multirow{3}{*}{Non-Commercial} & \multirow{3}{*}{\shortstack[l]{~~352\\\small{(19.0\%)}}} & \github & \cellcolor{colorIncorrect}{49} & \cellcolor{colorIncorrect}{303} & \cellcolor{colorCorrect}{0} & 0 \\
        & & \hf & \cellcolor{colorIncorrect}{113} & \cellcolor{colorIncorrect}{152} & \cellcolor{colorCorrect}{80} & 7 \\
        & & \pwc & \cellcolor{colorIncorrect}{2} & \cellcolor{colorIncorrect}{191} & \cellcolor{colorCorrect}{157} & 2 \\
        \midrule
        \multirow{3}{*}{Academic-Only} & \multirow{3}{*}{\shortstack[l]{~~~80\\\small{(4.3\%)}}} & \github & \cellcolor{colorIncorrect}{9} & \cellcolor{colorIncorrect}{71} & \cellcolor{colorIncorrect}{0} & \cellcolor{colorCorrect}{0} \\
        & & \hf & \cellcolor{colorIncorrect}{9} & \cellcolor{colorIncorrect}{65} & \cellcolor{colorIncorrect}{2} & \cellcolor{colorCorrect}{4} \\
        & & \pwc & \cellcolor{colorIncorrect}{5} & \cellcolor{colorIncorrect}{65} & \cellcolor{colorIncorrect}{2} & \cellcolor{colorCorrect}{8} \\
        
        \midrule
        \midrule
        \multirow{3}{*}{Total} & \multirow{3}{*}{\shortstack[l]{~~1858\\\small{(100\%)}}} & \github & 519 \footnotesize{(28\%)} & 1339 \footnotesize{(72\%)} & 0 \footnotesize{(0\%)} & 0 \footnotesize{(0\%)} \\
        & & \hf & 462 \footnotesize{(25\%)} & 1289 \footnotesize{(69\%)} & 89 \footnotesize{(5\%)} & 18 \footnotesize{(1\%)} \\
        & & \pwc & 351 \footnotesize{(19\%)} & 1299 \footnotesize{(70\%)} & 161 \footnotesize{(9\%)} & 47 \footnotesize{(3\%)} \\
        \bottomrule

    \end{tabular}
    \caption{\textbf{The distribution of license use categories shows our licenses have far fewer ``Unspecified'' omissions than GitHub (\github, 72\%), Hugging Face (\hf, 69\%), and Papers with Code (\pwc, 70\%), categorizing license more confidently into commercial or non-commercial categories.}
    GitHub, Hugging Face, and Papers with Code match our licenses (green regions) 43\%, 35\%, and 54\% of the time, respectively, and suggest incorrect licenses that are \textit{too permissive} 29\%, 27\%, and 16\% of the time.
    }
    \label{tab:license-dist}
\end{table*}

\subsection{Licenses in the Wild}
\label{sec:license-results}

This work constitutes the first extensive study of empirical license use for Natural Language Processing datasets. In this section, we share the insights we have gathered from our large-scale annotation and categorization.
There is an important assumption in this section: the OpenAI Terms of Use is a contract, not a license, which prohibits the development of competing models using its outputs.
For simplicity, we treat this as a Non-Commercial license in our analysis, though this is disputed for third parties who did not generate the OpenAI data themselves and therefore may not be bound by their terms (see ~\cref{llm-generation} for discussion).
Given the intention of OpenAI not to facilitate competitive commercial uses, we follow their categorization for this analysis.

\textbf{Frequency of license types}~\cref{fig:license-dist} shows the distribution of licenses. 
The most common licenses are CC-BY-SA 4.0 (15.7\%), the OpenAI Terms of Use (12.3\%), and CC-BY 4.0 (11.6\%).
While most licenses are common and recognizable, there is a long tail of variants with unique settings, as well as a large set of Custom licenses accounting for 9.6\% of all recorded licenses on their own.
\textbf{This wide license diversity illustrates the challenge to startups and less resourced organizations attempting to navigate responsible training data collection, its legality and ethics.}

\textbf{Distribution of Restrictive Licenses} In total, 85\% of dataset licenses request attribution, and 30\% include a share alike clause.\footnote{``Share alike'' is a copyright term meaning adaptations or copies of a work are required to be released under the same license as the original.}
Datasets which request attribution pose challenges for practitioners who commonly train on hundreds of datasets and either don't cite them at all \citep{openai2023gpt4, anil2023palm, touvron2023llama} or simply cite an aggregation of data, which often falls short of the license's conditions of attributing the specific repository or paper.
Futhermore, ``Share alike'' clauses poses challenges for practitioners re-packaging data collections usually with multiple conflicting share-alike licenses without a clear way to resolve them (like \citet{longpre2023flan, wang2022super} and others in the \dpic{}).
Frequently, practitioners will over-write share-alike licenses with more restrictive or even less restrictive conditions.

\textbf{Missing or Unspecified Licenses.} Next, we compare our manually reviewed licensing terms, to the licenses for the same datasets, as documented in the aggregators GitHub, HuggingFace, and Papers with Code.~\cref{tab:license-dist} shows that these crowdsourced aggregators have an extremely high proportion of missing (``Unspecified'') licenses, ranging from 69-72\%, as compared to our protocol which yields only 30\% ``Unspecified''. The problem with ``Unspecified'' licenses is that it is unclear whether it is due to a shortcoming of the aggregator or because creators intentionally released them without a license. Consequently, risk-averse developers are forced to avoid many valuable datasets, which they would use otherwise if they were given assurance that there is indeed no license. As part of~\dpic{}, we manually reassign 46-65\% of dataset licenses (depending on the platform), resulting in much higher coverage, thus giving risk-averse developers more confidence and breadth in their dataset utilization.

\textbf{Incorrectly Specified Licenses.}~\cref{tab:license-dist} also finds real licenses as assigned by us are frequently stricter than the ones by aggregators. GitHub, Hugging Face and Papers with Code each label license use cases too permissively in 29\%, 27\%, and 16\% of cases respectively. Our inspection suggests this is due to contributors on these platforms often mistaking licenses attached to code in GitHub repositories for licenses attached to data.

\definecolor{colorCorrect}{RGB}{227,244,221}
\definecolor{colorIncorrect}{RGB}{247, 220, 222}

\begin{table*}
    \centering
    \begin{tabular}{l|c|c|c|c|c|c}
        \toprule
        \textsc{Metrics} & \multicolumn{2}{c}{\textsc{Commercial}} &  \multicolumn{2}{c}{\textsc{Unspecified}} &  \multicolumn{2}{c}{\textsc{NC / A-O}} \\
        & \multicolumn{1}{c}{\textsc{Mean}} & \multicolumn{1}{c}{\textsc{Entropy}} & \multicolumn{1}{c}{\textsc{Mean}} & \multicolumn{1}{c}{\textsc{Entropy}} & \multicolumn{1}{c}{\textsc{Mean}} & \multicolumn{1}{c}{\textsc{Entropy}} \\
        \midrule

\textsc{Tasks} & 1.7\footnotesize{$\pm$0.1} & 0.61 & 1.6\footnotesize{$\pm$0.1} & 0.53 & \textbf{3.4}\footnotesize{$\pm$0.2} & \textbf{0.69} \\
\textsc{Languages} & \textbf{1.3}\footnotesize{$\pm$0.0} & \textbf{0.52} & 1.2\footnotesize{$\pm$0.0} & 0.16 & 1.1\footnotesize{$\pm$0.0} & 0.45 \\
\textsc{Topics} & 8.2\footnotesize{$\pm$0.2} & 0.70 & \textbf{9.2}\footnotesize{$\pm$0.1} & 0.75 & 9.1\footnotesize{$\pm$0.2} & \textbf{0.77} \\
\textsc{Sources} & 1.6\footnotesize{$\pm$0.1} & 0.67 & 1.8\footnotesize{$\pm$0.1} & 0.72 & \textbf{4.2}\footnotesize{$\pm$1.3} & \textbf{0.78} \\
\midrule
\textsc{Input Text Lengths} & \textbf{1043.4}\footnotesize{$\pm$151.9} & 6.37 & 860.2\footnotesize{$\pm$67.7} & \textbf{6.66} & 950.3\footnotesize{$\pm$112.9} & 6.46 \\
\textsc{Target Text Lengths} & 102.7\footnotesize{$\pm$14.6} & 4.39 & 90.5\footnotesize{$\pm$14.3} & 4.09 & \textbf{1580.7}\footnotesize{$\pm$965.6} & \textbf{5.37} \\
\textsc{Synthetic} & 12.8\%\footnotesize{$\pm$2.1} & - & 13.6\%\footnotesize{$\pm$1.7} & - & \textbf{45.5}\%\footnotesize{$\pm$3.4} & - \\
    \bottomrule
    \end{tabular}
    \caption{The mean number of features (e.g. tasks or languages) per dataset, and the mean entropy of the distribution, representing the diversity of categories.
    \textbf{Non-Commercial / Academic-Only datasets have consistently and statistically higher task, topic, and source variety than Commercial datasets.}
    We use Normalized Shannon Entropy for discrete features, and Differential Entropy for continuous features, which are both measures of randomness.
    }
    \label{tab:license-stats}
\end{table*}


\subsection{How does Data Availability Differ by License Use Category?}
\label{sec:license-characteristics}

While non-commercial and academic-only licenses play important roles in protecting data use, their presence can also exclude communities from participating (or competing) in the development of these technologies.
In this section, we break down datasets according to their license restrictions and see how they differ.
Specifically, we ask: \textit{Does complying with licenses dictate systematic differences in resources for commercially-permissive (``open'') and non-commercial (``closed'') development?} And what particular features of data are particularly constrained by non-commercial prohibitions?

We compare datasets by categories of permitted use, according to their licenses: (1) Commercially viable, (2) Non-Commercial/Academic-Only (\ncao), or (3) Unspecified license.
We group together Non-Commercial and Academic-Only conditions as the distinction will rarely matter for developers.
We argue in ~\cref{sec:legal-discussion} that datasets without any license (Unspecified) have not imposed any conditions, so can often be treated as commercially viable, but this may depend on a developer's risk tolerance and jurisdiction.

\begin{figure*}[htbp]
    \centering
    {{\includegraphics[width=0.99\textwidth]{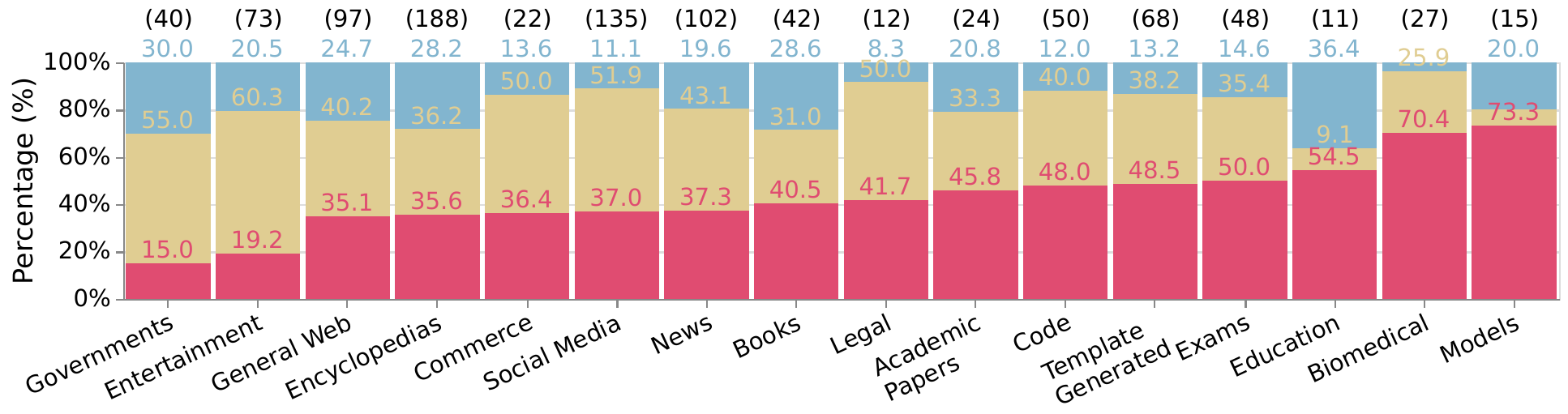}}} 
    \qquad
    {{\includegraphics[width=0.99\textwidth]{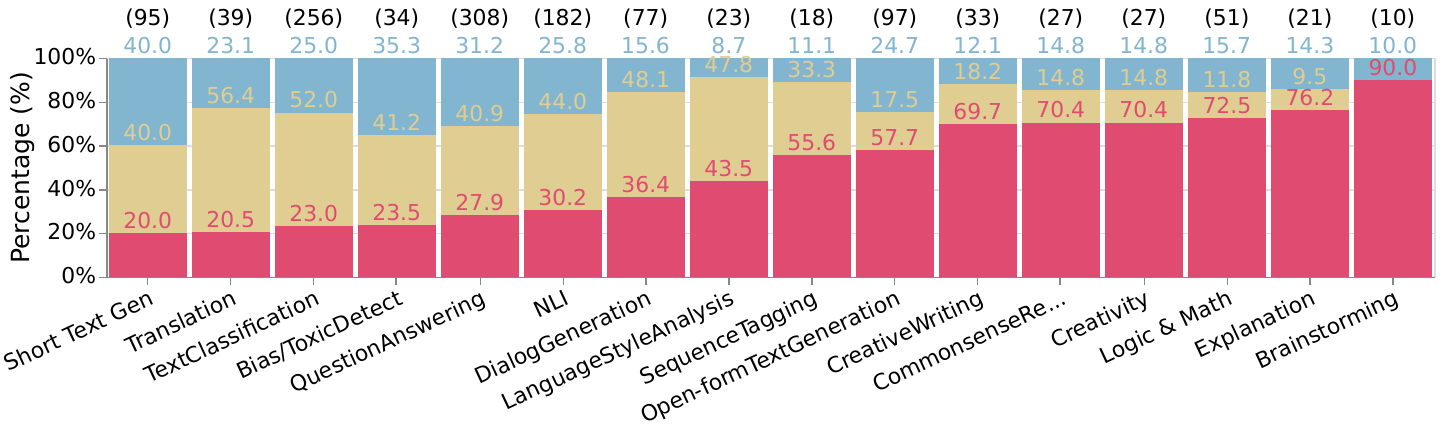}}}

    \caption{The distribution of datasets in each \textbf{Domain Source (top)} and \textbf{task (bottom)} category, with total count above the bars, and the portion in each license use category shown via bar color.
    \colorbox{ncred}{Red} is Non-commerical/Academic-Only, \colorbox{unspecgold}{Yellow} is Unspecified, and \colorbox{cblue}{Blue} is Commercial.
    \textbf{Creative, reasoning, and long-form generation tasks, as well as datasets sourced from models, exams, and the general web see the highest rate of non-commercial licensing.}}
    \label{fig:domain-task-licenses}
\end{figure*}

\textbf{Non-Commercial \& Academic-Only Licensed Datasets have statistically greater diversity in their representation of tasks, topics, sources, and target text lengths.}
For each of these features,~\cref{tab:license-stats} illustrates the mean number per dataset, broken down by license category and entropy to measure the randomness, and thus diversity, of each feature.
~\ncao{} datasets see greater diversity of tasks, topics, and sources represented in the text than commercial datasets.
~\cref{fig:domain-task-licenses} shows where this diversity comes from.
The most~\ncao{} task categories include Brainstorming, Explanation, Logic \& Math, as well as Creativity and Creative Writing.
In comparison, the most commercially viable task categories are Short Text Generation, Translation, and Classification.
Similarly, among Source Domains, Governments and Search Queries are largely viable for commercial (and unspecified) purposes, whereas General Web, Exams, and Model-generated sources are among the most restrictive.

\begin{figure*}[htbp]
    \centering
    {{\includegraphics[width=0.99\textwidth]{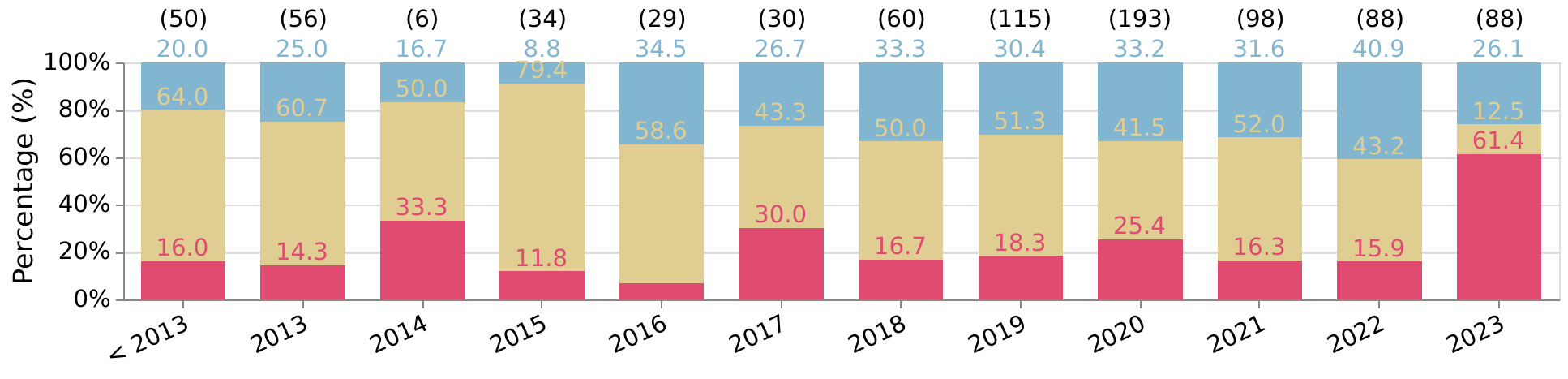}}}
    \qquad
    {{\includegraphics[width=0.99\textwidth]{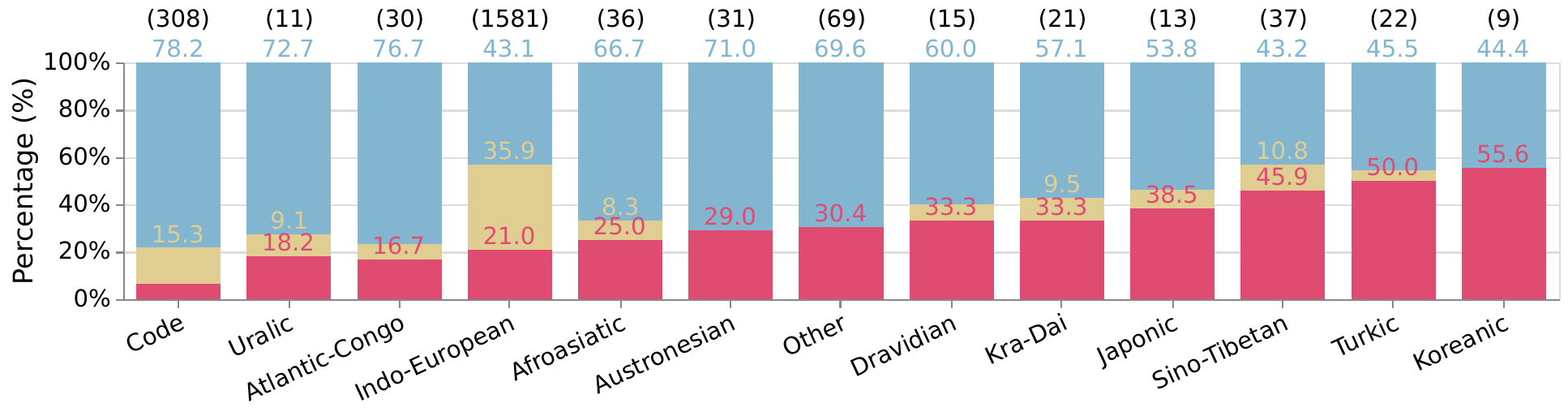}}}
    \caption{The distribution of datasets in each time of collection (top) and language family (bottom) category, with total count above the bars, and the portion in each license use category shown via bar color.
    \colorbox{ncred}{Red} is Non-commerical/Academic-Only, \colorbox{unspecgold}{Yellow} is Unspecified, and \colorbox{cblue}{Blue} is Commercial.
    \textbf{Lower resource languages, and datasets created in 2023 see a spike in non-commercial licensing.}
    }
    \label{fig:time-lang-licenses}
\end{figure*}

\begin{figure*}[htbp]
    \centering
    \subfloat[\centering \textbf{License Use Categories vs Text Lengths}\label{fig:license-scatter}]{{\includegraphics[width=0.5\textwidth]{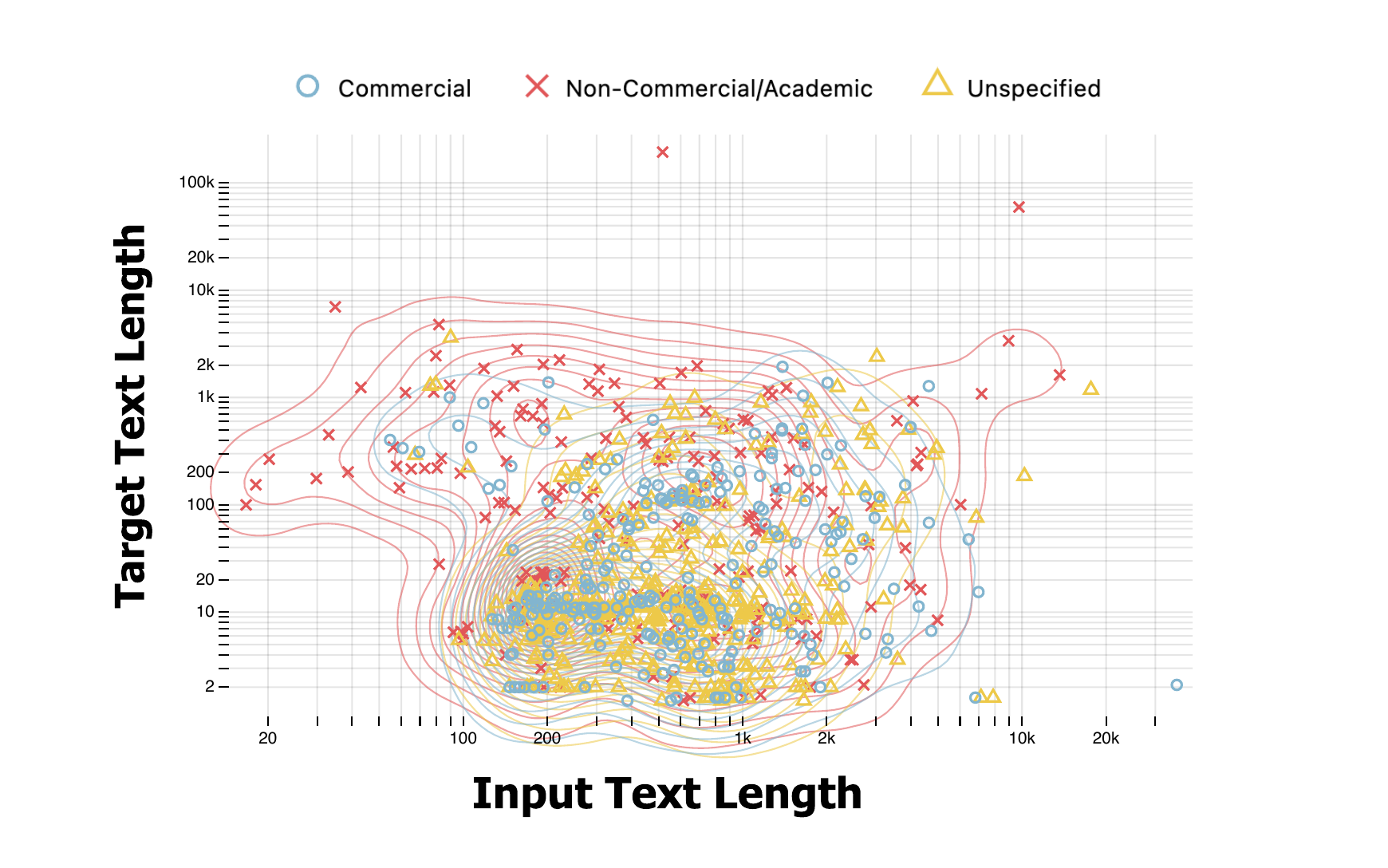}}}
    \subfloat[\centering \textbf{Synthetic/Regular Datasets vs Text Lengths}\label{fig:synthetic-scatter}]{{\includegraphics[width=0.5\textwidth]{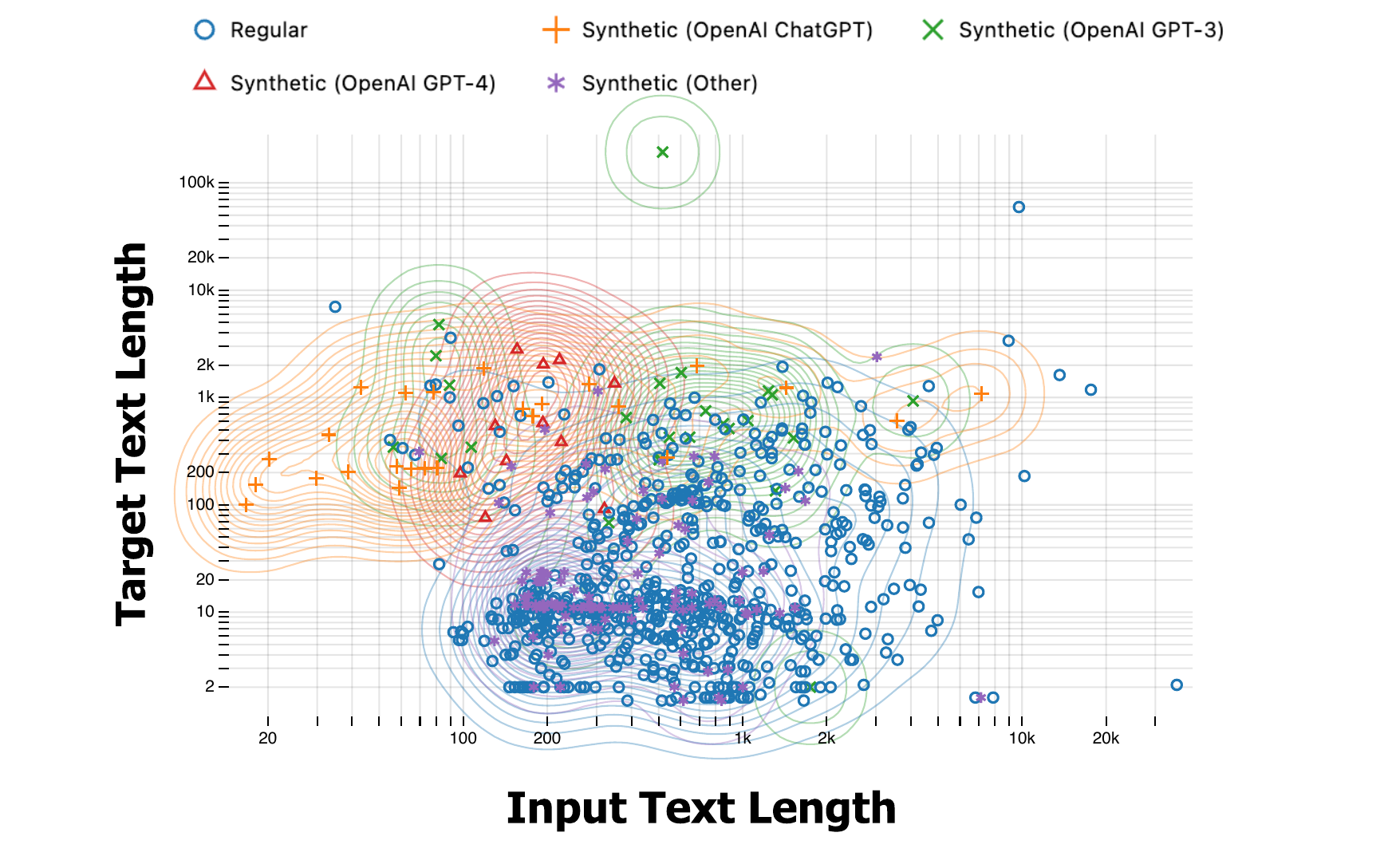}}}
    \caption{Across finetuning datasets, we visualize their mean input (x-axis) and target (y-axis) text lengths, measured in log-scaled number of words. 
    The colors indicate either their license use category (left) or whether they were machine generated or human collected (right).
    \textbf{Long target texts are represented in large part by Non-Commercial and Synthetic datasets, that are often generated by commercial APIs.}}
    \label{fig:text-len-characteristics}
    \vspace{-2mm}
\end{figure*}

\textbf{Target Text Lengths are significantly higher for~\ncao{} datasets than commercial datasets.}
Not only do~\ncao{} datasets appear more textually and functionally diverse, their length characteristics differ substantially.
While~\cref{tab:license-stats} shows the input text lengths across license categories are similar on average, the target text lengths are significantly higher for~\ncao{} datasets (103 vs 677).
This breakdown is further illustrated in~\cref{fig:text-len-characteristics}, where we see greater representation of both~\ncao{} and synthetic datasets above the 100 target token threshold (y-axis).

\textbf{The rise of synthetic datasets generated using APIs with non-commercial terms of use may explain the differences in text diversity and length.}
~\cref{tab:license-stats} also shows a full 45\% of~\ncao{} datasets are synthetic, as compared to $<14\%$ in more permissive license categories.
~\citet{alpaca,selfinstruct2022,xu2023wizardlm} and their variants, all generated in part using commercial APIs, exhibit stronger task and topic diversity than traditional academic datasets, as they cater to longer form generations, by design.
This is evident from the concentration of creative, brainstorming, and reasoning tasks baked into them, as compared to the focus of more topic-focused question answering, classification, and short text generation in non-synthetic datasets.
These datasets are usually created using larger proprietary models, mostly from OpenAI APIs.
The OpenAI Terms of Use state ``you may not...use output from the Services to develop models that compete with OpenAI.'' which we discuss in~\cref{sec:legal-discussion}.\footnote{\url{https://openai.com/policies/terms-of-use}}

\textbf{2023 has a large spike in license usage, and in ~\ncao{} licensed data, representing 61\%, as compared to 20\% on average in prior years.}
Among the large collection of datasets we trace, we record the date at which they are released, by cross-referencing their associated GitHub, ArXiv, and Hugging Face dates.
We find a striking change in the pattern of licensing restrictions.
As shown in~\cref{fig:time-lang-licenses}, prior to 2023, no year saw greater than 1/3 of the datasets released as~\ncao{}.
However, in 2023, which includes many of the most popular and diverse datasets, the~\ncao{} rate is 61\%.
Furthermore, most datasets were unaccompanied by a license prior to 2022 (\~50-80\%), as compared to only 12\% in 2023.
The shift to more license use, and more restrictively conditioned data releases may foretell future challenges to open data, if the trend continues.

\begin{figure}[ht]
\centering
    \includegraphics[width=0.99\textwidth]{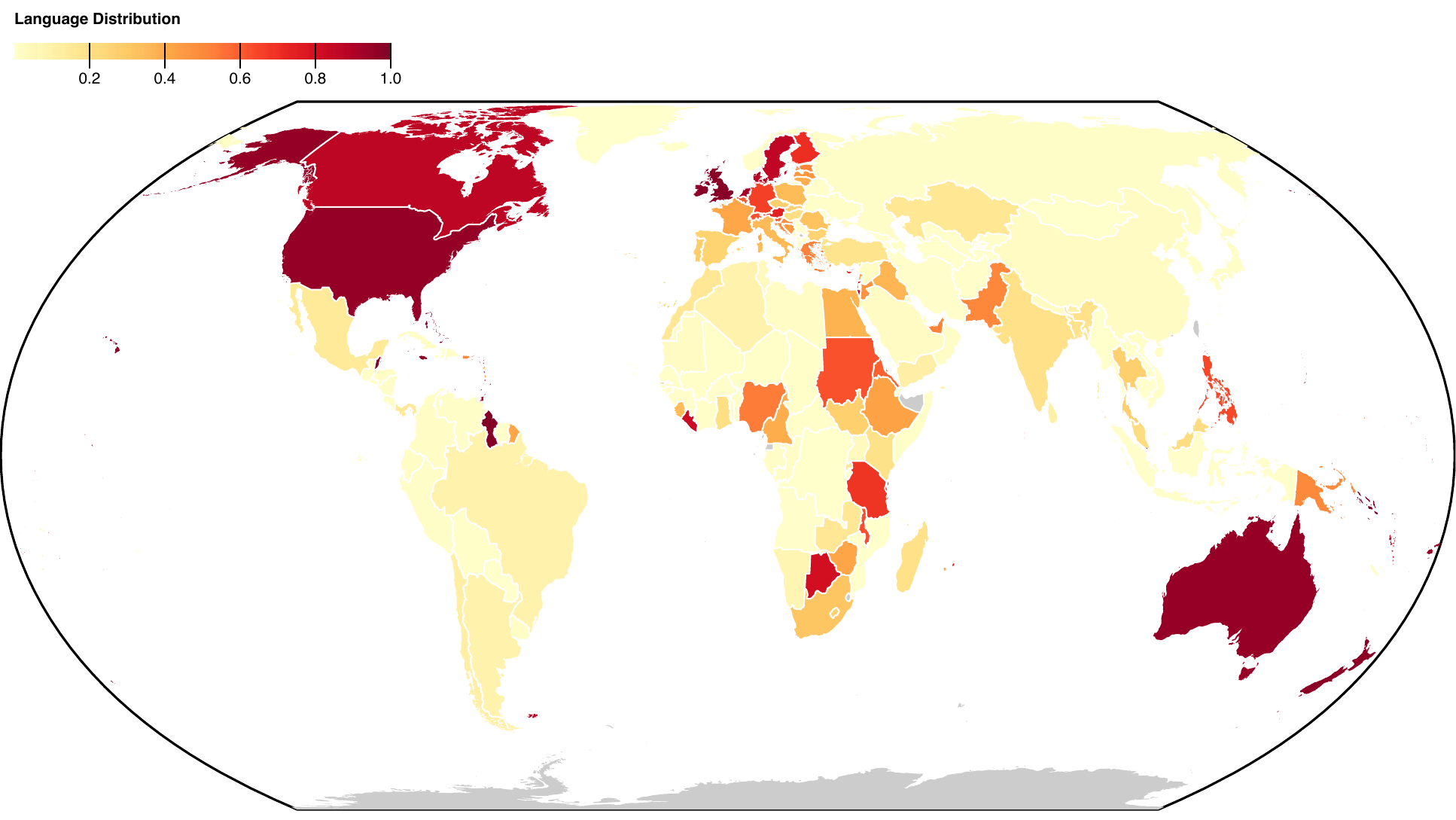}
    \caption{A global heatmap measuring how well each country's spoken languages are represented by the composition of natural language datasets in ~\dpic{}, as calculated by ~\cref{eq:lang-rep}.
    \textbf{English-speaking and Western European nations are best represented, while the Global South sees limited coverage.}}
    \label{fig:lang-map}
\vspace{-3mm}
\end{figure}

\textbf{Commercial datasets have greater language variety, but low-resource language datasets see the least commercial coverage.}
~\cref{tab:license-stats} shows that commercial datasets actually have greater diversity of languages than~\ncao{}.
However, when broken down by language family, as in~\cref{fig:time-lang-licenses}, we see stark differences in permitted use by group.
Code language datasets are nearly all commercially viable (78\%), because dataset creators can easily filter GitHub for permissively licensed repositories. 
Interestingly, English, Atlantic-Congo, and Afroasiatic languages also see large permissive representation.
However, Turkic, Sino-Tibetan, Japonic, and Indo-European languages see in excess of 35\% as non-commercial.
Note that while the Indo-European language family contains many high-resource European language families, there is a long tail of lower-resource ones.
These~\ncao{} language families provide directions for open data practitioners to focus their future efforts.

\subsection{Broader Characteristics of the Data}
\label{sec:dataset-characteristics}

In addition to understanding systematic differences in the data by license, there are research questions regarding the overall composition and characteristics of these widely used and adopted datasets.
Our compilation of metadata through the~\dpic{} allows us to map the landscape of data characteristics, and inspect particular features.
Note that all these details are also available with interactive visualizations at \url{www.comingsoon.com}, for further research and examination.

\textbf{Language representation is heavily skewed to English and Western European Languages.}
Following~\citet{talat2022you}'s recommendations in data transparency and documentation in demographic analysis, and corroborating~\citet{kreutzer2022quality}'s similar analysis for pretraining corpora, we find a stark Western-centric skew in representation.
~\cref{fig:lang-map} illustrates the coverage per country according to the spoken languages and their representation in \dpic{}. 
We compute a Language Representation score $S_k$ for each country $k$, parametrized by \( p_{kl} \), the percentage of people in country \( k \) that speak language \( l \), and \( w_{li} \) which is a binary indicator that is 1 if dataset \( i \in D \) contains language \( l \) and 0 otherwise.

$$S_k = \sum_{l \in L} \left( p_{kl} \times \sum_{i \in D} w_{li} \right)$$
\label{eq:lang-rep}

The distribution visualized in~\cref{fig:lang-map} shows that Asian, African, and South American nations are sparsely covered if at all.
Even when nations from the Global South appear to have linguistic representation, according to~\cref{eq:lang-rep}, the text source and dialect of the language contained in these datasets almost always originates from North American or European creators and web sources (though this is difficult to measure precisely). 
These observations corroborate similar findings in the geo-diversity of image data in the vision domain~\citep{shankar2017no, de2019does,mahadev2021understanding}.
The resulting models trained on these datasets are likely to have inherent bias, underperforming in critical ways for users of models outside of the west \citep{ahia-etal-2021-low-resource}. 

\begin{table*}[htbp]
    \centering
    \subfloat[\centering \textbf{Creators}\label{tab:creators}]{{\begin{tabular}{l|c}
\toprule
\multicolumn{1}{c}{\textsc{Name}} & \textsc{Pct} \\
    \midrule
\multicolumn{1}{c}{\textsc{Academic}} & 68.7\% \\
\midrule
\footnotesize{University of Washington} & \footnotesize{8.9\%} \\
\footnotesize{Stanford University} & \footnotesize{6.8\%} \\
\footnotesize{New York University} & \footnotesize{5.4\%} \\
\footnotesize{University of Southern...} & \footnotesize{3.5\%} \\
\footnotesize{Carnegie Mellon Univer...} & \footnotesize{3.5\%} \\
\footnotesize{Saarland University} & \footnotesize{2.6\%} \\
\footnotesize{Cardiff University} & \footnotesize{2.3\%} \\
\midrule
\multicolumn{1}{c}{\textsc{Industry Lab}} & 21.4\% \\
\midrule
\footnotesize{Facebook AI Research} & \footnotesize{8.4\%} \\
\footnotesize{Microsoft Research} & \footnotesize{4.1\%} \\
\footnotesize{Google Research} & \footnotesize{2.9\%} \\
\footnotesize{DeepMind} & \footnotesize{1.9\%} \\
\footnotesize{Microsoft Semantic Mac...} & \footnotesize{0.9\%} \\
\footnotesize{NAVER AI Lab} & \footnotesize{0.8\%} \\
\footnotesize{Salesforce Research} & \footnotesize{0.7\%} \\
\midrule
\multicolumn{1}{c}{\textsc{Research Group}} & 17.1\% \\
\midrule
\footnotesize{AI2} & \footnotesize{12.3\%} \\
\footnotesize{CLUE team} & \footnotesize{0.5\%} \\
\footnotesize{Alan Turing Institute} & \footnotesize{0.5\%} \\
\footnotesize{CodeX} & \footnotesize{0.4\%} \\
\footnotesize{Qatar Computing Resear...} & \footnotesize{0.4\%} \\
\footnotesize{Barcelona Supercomputi...} & \footnotesize{0.4\%} \\
\footnotesize{BigCode} & \footnotesize{0.2\%} \\
\midrule
\multicolumn{1}{c}{\textsc{Corporation}} & 15.8\% \\
\midrule
\footnotesize{Google} & \footnotesize{2.1\%} \\
\footnotesize{IBM} & \footnotesize{2.0\%} \\
\footnotesize{Microsoft} & \footnotesize{1.4\%} \\
\footnotesize{Wind Information Co.} & \footnotesize{1.4\%} \\
\footnotesize{Snap Inc.} & \footnotesize{1.3\%} \\
\footnotesize{Meta} & \footnotesize{1.1\%} \\
\footnotesize{Synapse Développement} & \footnotesize{1.1\%} \\
\midrule
\multicolumn{1}{c}{\textsc{Startup}} & 4.0\% \\
\midrule
\footnotesize{OpenAI} & \footnotesize{1.3\%} \\
\footnotesize{NomicAI} & \footnotesize{0.8\%} \\
\footnotesize{Omniscien Technologies} & \footnotesize{0.4\%} \\
\footnotesize{Anthropic AI} & \footnotesize{0.2\%} \\
\footnotesize{EightSleep} & \footnotesize{0.2\%} \\
\footnotesize{Curai} & \footnotesize{0.2\%} \\
\footnotesize{IMRSV Data Labs} & \footnotesize{0.2\%} \\
\midrule
\multicolumn{1}{c}{\textsc{Other}} & 0.7\% \\
\bottomrule
\end{tabular}}}
    \hspace{0.5cm}
    \subfloat[\centering \textbf{Topics}\label{tab:topics}]{{\begin{tabular}{l|c}
\toprule
\multicolumn{1}{c}{\textsc{Name}} & \textsc{Pct} \\
    \midrule
\multicolumn{1}{c}{\textsc{Question Answering}} & 36.0\% \\
\midrule
\footnotesize{Question Answering} & \footnotesize{27.7\%} \\
\footnotesize{Multiple Choice Questi...} & \footnotesize{3.9\%} \\
\footnotesize{Information Extraction} & \footnotesize{1.8\%} \\
\midrule
\multicolumn{1}{c}{\textsc{Text Classification}} & 29.9\% \\
\midrule
\footnotesize{Text Classification} & \footnotesize{16.1\%} \\
\footnotesize{Sentiment Analysis} & \footnotesize{9.8\%} \\
\footnotesize{Named Entity Recognition} & \footnotesize{4.3\%} \\
\midrule
\multicolumn{1}{c}{\textsc{Natural Language Inf...}} & 21.1\% \\
\midrule
\footnotesize{Textual Entailment} & \footnotesize{14.6\%} \\
\footnotesize{Natural Language Infer...} & \footnotesize{5.3\%} \\
\footnotesize{Fact Verification} & \footnotesize{1.3\%} \\
\midrule
\multicolumn{1}{c}{\textsc{Open-form Text Gener...}} & 11.3\% \\
\midrule
\footnotesize{Open-form Text Generation} & \footnotesize{2.2\%} \\
\footnotesize{Title Generation} & \footnotesize{1.5\%} \\
\footnotesize{Inverted Summarization} & \footnotesize{1.2\%} \\
\midrule
\multicolumn{1}{c}{\textsc{Short Text Generation}} & 10.9\% \\
\midrule
\footnotesize{Question Generation} & \footnotesize{4.0\%} \\
\footnotesize{Fill in The Blank} & \footnotesize{1.4\%} \\
\footnotesize{Inverted Multiple-Choi...} & \footnotesize{0.9\%} \\
\midrule
\multicolumn{1}{c}{\textsc{Dialog Generation}} & 9.0\% \\
\midrule
\footnotesize{Dialogue Generation} & \footnotesize{4.2\%} \\
\footnotesize{Dialog Generation} & \footnotesize{3.7\%} \\
\footnotesize{Dialogue Act Recognition} & \footnotesize{0.4\%} \\
\midrule
\multicolumn{1}{c}{\textsc{Summarization}} & 6.3\% \\
\midrule
\footnotesize{Summarization} & \footnotesize{5.7\%} \\
\footnotesize{Simplification} & \footnotesize{0.5\%} \\
\footnotesize{Summarization of US Co...} & \footnotesize{0.1\%} \\
\midrule
\multicolumn{1}{c}{\textsc{Logical and Mathemat...}} & 6.0\% \\
\midrule
\footnotesize{Logical Reasoning} & \footnotesize{2.3\%} \\
\footnotesize{Data Analysis} & \footnotesize{2.0\%} \\
\footnotesize{Algebraic Expression E...} & \footnotesize{1.2\%} \\
\midrule
\multicolumn{1}{c}{\textsc{Code}} & 4.8\% \\
\midrule
\multicolumn{1}{c}{\textsc{Response Ranking}} & 4.4\% \\
\midrule
\multicolumn{1}{c}{\textsc{Translation}} & 4.4\% \\
\midrule
\multicolumn{1}{c}{\textsc{Creative Writing}} & 3.9\% \\
\midrule
\multicolumn{1}{c}{\textsc{Other}} & 23.9\% \\
\bottomrule
\end{tabular}}}
    \hspace{0.5cm}
    \subfloat[\centering \textbf{Domains \& Sources}\label{tab:domains}]{{\begin{tabular}{l|c}
\toprule
\multicolumn{1}{c}{\textsc{Name}} & \textsc{Pct} \\
    \midrule
\multicolumn{1}{c}{\textsc{Encyclopedias}} & 21.5\% \\
\midrule
\footnotesize{wikipedia.org} & \footnotesize{14.6\%} \\
\footnotesize{wikihow.com} & \footnotesize{2.7\%} \\
\footnotesize{dbpedia} & \footnotesize{1.4\%} \\
\midrule
\multicolumn{1}{c}{\textsc{Social Media}} & 15.9\% \\
\midrule
\footnotesize{reddit} & \footnotesize{6.2\%} \\
\footnotesize{twitter} & \footnotesize{4.0\%} \\
\footnotesize{quora} & \footnotesize{1.6\%} \\
\midrule
\multicolumn{1}{c}{\textsc{General Web}} & 11.2\% \\
\midrule
\footnotesize{undisclosed web} & \footnotesize{7.0\%} \\
\footnotesize{commoncrawl.org} & \footnotesize{2.5\%} \\
\footnotesize{data.world/samayo/coun...} & \footnotesize{0.6\%} \\
\midrule
\multicolumn{1}{c}{\textsc{News}} & 11.1\% \\
\midrule
\footnotesize{cnn.com} & \footnotesize{1.6\%} \\
\footnotesize{financial news} & \footnotesize{1.5\%} \\
\footnotesize{press releases} & \footnotesize{1.4\%} \\
\midrule
\multicolumn{1}{c}{\textsc{Entertainment}} & 8.5\% \\
\midrule
\footnotesize{opensubtitles.org} & \footnotesize{2.5\%} \\
\footnotesize{imdb.com} & \footnotesize{1.6\%} \\
\footnotesize{travel guides} & \footnotesize{1.3\%} \\
\midrule
\multicolumn{1}{c}{\textsc{Code}} & 5.7\% \\
\midrule
\footnotesize{stackexchange.com} & \footnotesize{2.0\%} \\
\footnotesize{github} & \footnotesize{1.2\%} \\
\footnotesize{opus software projects} & \footnotesize{0.9\%} \\
\midrule
\multicolumn{1}{c}{\textsc{Exams}} & 5.6\% \\
\midrule
\footnotesize{web exams} & \footnotesize{2.9\%} \\
\footnotesize{gmat} & \footnotesize{1.1\%} \\
\footnotesize{gre exams} & \footnotesize{0.9\%} \\
\midrule
\multicolumn{1}{c}{\textsc{Books}} & 4.9\% \\
\midrule
\footnotesize{project gutenberg} & \footnotesize{2.0\%} \\
\footnotesize{non-fiction books} & \footnotesize{1.3\%} \\
\footnotesize{fiction books} & \footnotesize{1.3\%} \\
\midrule
\multicolumn{1}{c}{\textsc{Governments}} & 4.7\% \\
\midrule
\multicolumn{1}{c}{\textsc{Biomedical}} & 3.2\% \\
\midrule
\multicolumn{1}{c}{\textsc{Search Queries}} & 3.0\% \\
\midrule
\multicolumn{1}{c}{\textsc{Academic Papers}} & 2.8\% \\
\midrule
\multicolumn{1}{c}{\textsc{Other}} & 61.2\% \\
\bottomrule
\end{tabular}}}
    \caption{A summary of the distribution of \textbf{Creators}, \textbf{Topics}, and \textbf{Source Domains} across all 1800+ datasets. Datasts can have multiple creators, text topics, and sources.}
    \label{tab:distributions}
\end{table*}


\textbf{The primary drivers of dataset curation are Academic organizations, supplying 69\%, followed by 21\% industry labs, and 17\% research institutions.} 
These metrics describe the scale of dataset curation contributions, but not the influence each dataset has had on the community.
~\cref{tab:creators} demonstrates the single largest dataset contributors are AI2 (12.3\%), University of Washington (8.9\%), and Facebook AI Research (8.4\%). It is important to note that these contributors often only download and compile text from the Internet that was originally written by other people.

\textbf{Text datasets focus on topics of Language \& Linguistics, General Knowledge, Logic, \& Lifestyle.}
Prior data collection work focuses predominantly on describing datasets by their task compositions \citep{sanh2021multitask,selfinstruct2022,longpre2023flan}, but rarely by their actual topics (except~\citep{gao2020pile} in their Appendix).~\cref{tab:topics} shows the most popular topics, clustered by category, with their representation across datasets.
Like most NLP tasks, much of this text data focuses on communication and language understanding topics, followed closely by general knowledge, routine, sports, and education.

\textbf{Text datasets are sourced primarily from Online Encyclopedias (22\%), Social Media (16\%), scraped from the General Web (11\%), News (11\%), Entertainment web resources (9\%).}
While practitioners document their individual dataset sources in their published papers, this information is unstructured and can be hard to find. 
As a result, massive collections of widely used datasets rarely compile the distribution of their original sources, instead just citing the papers.
After a series of dataset compilations and re-packaging, the original sources are often lost or not well known.
By manually scanning approximately 500 academic papers our volunteers annotated the original text sources and compiled them into domain clusters, to permit attribution and analysis, as summarized in~\cref{tab:domains}.
Among the individual most adopted sources by the used sources are wikipedia.org (14.9\%), undisclosed webpage scrapes (7.0\%), reddit (6.2\%), and Twitter (4.0\%). The least represented domains are Commerce, Reviews, Legal, Academic Papers, and Search Queries, among others.

\vspace{-2mm}
\section{Legal Discussion}
\label{sec:legal-discussion}

Our empirical analysis highlights that we are in the midst of a crisis in dataset provenance and practitioners are forced to make decisions based on limited information and opaque legal frameworks. 
While we believe our tooling will enable better transparency about where licenses are in tension, major legal ambiguities remain in data licensing. 

\vspace{-2mm}
\paragraph{Background} Copyright laws aim to encourage written and artistic expression by giving authors exclusive rights to copy, distribute, and adapt their work~\citep{patterson2003copyright, burger1988berne}.
Open-source licenses first emerged as legal tools to encourage collaboration around software development~\citep{von2003special}.
A range of licenses with different terms and purposes exists including the MIT License, Creative Commons Licenses, and the Apache License, as well as the newer Responsible AI License (RAIL) and AI2 ImpACT Licenses.\footnote{See \url{https://www.licenses.ai/blog/2023/3/3/ai-pubs-rail-licenses} and \url{https://allenai.org/impact-license\#licenses}. These license templates propose terms aimed at encouraging more responsible or risk-based machine learning practices, see also \cite{contractor2022behavioral}}
The interplay between copyright and licenses can be understood in the following way: copyright automatically gives creators exclusive rights in their works and creators assign these rights to others through license agreements.
As we will explore, the open-source licenses that emerged in the last three decades are not always well-equipped to handle the unique characteristics of data, and especially supervised AI training data. 
Meanwhile, it remains unclear how relevant laws, including those related to copyright and fair use, should be applied to the unique challenges raised by Generative AI and supervised datasets~\citep{lee2023talkin}.
In this section, we highlight some of the key legal challenges and ambiguities related to supervised datasets.

\vspace{-2mm}
\paragraph{Lifecycle of a dataset} 
We focus on \emph{supervised datasets}, which we define as datasets that are created for machine learning (mainly for finetuning and alignment) and where dataset creators made copyrightable contributions in the form of annotations or compilations. 
A typical supervised dataset is the result of a process that involves several stages of scraping (or machine generation) and annotation by different entities. Generally, raw data is created by people interacting with internet platforms, such as individuals writing articles, sharing artworks, or engaging in online discussion forums.
The copyrights to this raw data are normally held by individual users (e.g. Reddit) or by the platform (e.g. Amazon Reviews).
Much of this data has been scraped to construct unsupervised datasets for machine learning and this use is commonly justified on the basis of fair use or data mining exceptions to copyright~\citep{henderson2023foundation, sobel2017artificial, lee2023talkin, samuelson2023generative, lemley2020fair}.
However, we find that many common supervised datasets are generated by annotating small samples of scraped raw data using human annotators or large language models.
The annotated data is then published with a license agreement.
In stark contrast to the copyrighted content that is scraped from the web, supervised datasets were created for the sole purpose of furthering machine learning.
The focus of the legal discussion in this section is on how supervised dataset creators can constrain the usage of the copyrightable content they create through licenses and other legal mechanisms.
Though we do not address them here, there are several important related questions on the use of copyrighted works to create supervised datasets and on the copyrightability of training datasets.

\begin{tcolorbox}[
  colframe=mydarkblue,
  coltitle=white,
  fonttitle=\bfseries,
  title=Surpervised Dataset Example: SQuAD,
  sharp corners=south,
  rounded corners=north,
  width=\textwidth
]
  \cite{rajpurkar2016squad} present a prototypical supervised dataset on reading comprehension. To create the dataset, the authors take paragraph-long excerpts from 539 popular Wikipedia articles and hire crowd-source workers to generate over 100,000 questions whose answers are contained in the excerpt. For example:\\
  
  \textbf{Wikipedia Excerpt} \textit{In meteorology, precipitation is any product of the condensation of atmospheric water vapor that falls under gravity.} \\
  \textbf{Worker-generated question:} What causes precipitation to fall? \textbf{Answer:} Gravity \\

  Here the authors use Wikipedia text as a basis for their data and their dataset contains 100,000 new question-answer pairs based on these texts.  
\end{tcolorbox}

\vspace{-2mm}
\paragraph{Copyright laws vary by jurisdiction and are subjective, so it is challenging to develop technical safeguards that guarantee compliance.}
The legal analysis surrounding supervised datasets is complicated by the lack of a uniform global legal framework to address copyright concerns. 
Different jurisdictions have different and evolving laws.
Therefore, the location of model developers and training data creators as well as where and when data was collected may influence the legal analysis.
For example, the United States has a fair-use exception to copyright that allows the limited use of copyrighted material under certain circumstances without requiring permission from the rights holders (17 U.S.C. \S107) .
The EU has no fair-use provision but does have an explicit copyright exception to allow data mining under certain conditions, like obtaining lawful access to the data~\citep{margoni2022deeper}.
Meanwhile, datasets themselves generally enjoy copyright protection in the U.S.~\citep{lee2023talkin} while the E.U. recently created a unique set of rights for dataset creators with the purpose of incentivizing research and development related to databases~\citep{derclaye2022sui}. 
In addition to differences across jurisdictions, there are also several international agreements related to copyright~\cite{ricketson2022international}.
Ultimately, it can be challenging to determine which laws should apply to a given machine learning project when the relevant rules vary between the locations where the data was scraped and annotated, where it was downloaded, where the model was trained, and where the model was deployed.

While geographical disparities in regulatory frameworks present one set of challenges, the subjectivity inherent in determining whether copyright infringement has occurred makes it even more challenging to design technical safeguards.
For example, in the U.S. part of the copyright infringement analysis depends on whether two works are subjectively similar from the perspective of an ordinary person~\citep{mohler1999toward, cohen1986masking, balganesh2014judging}.
This is a subjective standard and existing case law may be challenging to extend to generative AI outputs. 
As a result, while there are technical strategies that can reduce the risk of infringement~\citep{henderson2023foundation, sag2023copyright, vyas2023provable}, it will be difficult for developers to create technical safeguards that eliminate this risk entirely.

\vspace{-2mm}
\paragraph{Open legal question regarding copyright and model training.}
Apart from these jurisdictional and interpretive ambiguities, the process of training a model raises specific copyright questions~\citep{epstein2023art}.
Training a model poses several interesting legal questions with respect to copyright and infringement may occur in several ways even before any outputs are generated.

First, the act of creating a training dataset by scraping existing works involves making a digital copy of the underlying data. As the name implies, copyright gives the author of a protected work the exclusive right to make copies of that work.
If the scraped data is protected by copyright, then creating training data corpora may raise copyright issues~\citep{quang2021does}.
Second, copyright holders generally have an exclusive right to create derivative works (e.g., translations of a work) but it is not clear whether a trained machine learning model should be considered a derivative of the training data~\citep{lee2023talkin}.
If models are considered to be derivative works, then training a model would be more likely to violate the rights of the training data's copyright holders~\citep{gervais2021ai}.

In the U.S., the fair use exception may allow models to be trained on protected works~\citep{henderson2023foundation,lemley2020fair,sobel2017artificial, samuelson2023generative}.
As these authors explain, the training of machine learning models on copyrighted content may be permissible if the underlying works are significantly ``transformed'' into model weights, only a small amount of each work in the training data is included in the trained model, model training is designed to only glean generalizable insights from the training data, and the trained model does not have a strong effect on the economic success of the works in the training data.
It is important to underscore that, while training a machine learning model itself may be protected by fair use this does not mean that model outputs will not infringe on the copyright of prior works.
As the authors above highlight, the application of fair use in the context is still evolving and several of these issues are currently being litigated (see e.g., \textit{Andersen v. Stability}, \textit{Doe v. GitHub}, and \textit{Tremblay v. OpenAI}).

\vspace{-2mm}
\paragraph{Fair use is less likely to apply when works are created for the sole purpose of training machine learning models as in the case of supervised datasets with copyrightable compositions or annotations.}
The prior literature on fair use and machine learning tends to focus on copyrighted art or text that was scraped to train a model.
These scraped works were not created for the purpose of training machine learning models.
By contrast, in this paper, we focus on supervised datasets that were created for the sole purpose of training machine learning models.
As underscored by \cite{henderson2023foundation} and \cite{sobel2017artificial}, the fair use analysis depends in part on whether a trained model copies the ``expressive purpose'' of the original work.
While the expressive purpose of a piece of text or art is not to train machine learning models, the purpose of a training dataset is to do just that.
As a result, we expect that it is less likely that fair use would apply to the use of curated data.
Instead, the creators of these datasets hold a copyright in the dataset\footnote{Data ownership and data copyright are complex topics~\citep{ginsburg1992no}. 
We assume that the creators of supervised datasets have some form of copyright in their dataset, though there is often content in these datasets that is owned by third parties.
If they satisfy the requirements for copyrightability, dataset creators would have a copyright interest in any new content they create (e.g. annotations).
In the U.S., datasets themselves may also be copyrightable as compilations~\citep{lee2023talkin} while the E.U. provides so-called \textit{sui generis} rights for databases~\citep{derclaye2022sui}.} and the terms of the dataset license agreement govern the subsequent use of this data 
However, it is rare in practice for an LLM to use a single supervised dataset and often multiple datasets are compiled into collections.
This further complicates the legal analysis because we find that the license terms of many popular dataset collections are conflicting.

\vspace{-2mm}
\paragraph{Licenses used for datasets are often ill-suited for this purpose.}
Beyond the intricate interplay between training data and fair use, the frequently misapplied licensing frameworks for datasets present another set of complications. 
Most open-source licenses were designed for software, but we find them being attached to datasets.
These licenses were intended to be applied to software, not data, which creates challenges~\citep{meeker2022beyond}.
One of the challenges is that licenses like the Apache and the Creative Commons outline restrictions related to ``derivative'' or ``adapted works'' but it remains unclear if a trained model should be classified as a derivative work. 
This issue is further exacerbated when multiple datasets, each potentially governed by a different open-source license, are amalgamated into collections. 
If the requirements of the underlying license agreements are irreconcilable, such as different copyleft requirements, this makes it extremely hard for developers to use certain collections while respecting all license terms.
To remedy these issues, new licenses are being proposed to address the needs of machine learning datasets such as the BigScience Responsible AI License or an adaptation of the MIT License that requires additional permissions for model training proposed by \cite{ioannidis2023chatgpt}.
Despite these new proposals, we find that the majority of datasets are licensed under conventional open-source licenses.

\vspace{-2mm}
\paragraph{LLM-generated annotations raise additional legal considerations}\label{llm-generation}
We find that approximately 12\% of the datasets we audit were annotated using OpenAI. 
The OpenAI Terms of Use state that outputs from the OpenAI service may not be used to ``to develop models that compete with OpenAI''\footnote{\texttt{https://openai.com/policies/terms-of-use}}. 
These terms seem to preclude a developer from using OpenAI to generate training data to train a competing LLM.
However, it is not clear whether they would also limit the ability of a developer to use OpenAI to create and publish an annotated dataset.
On the one hand, publishing such a dataset does not directly compete with OpenAI.
On the other hand, it seems foreseeable that such a dataset could enable third parties (who did not themselves use OpenAI) to create competing LLMs.
In the U.S., there are several doctrines of secondary or indirect copyright liability aimed to enforce copyright in cases where there is no direct infringement~\citep{grossman2005sony, lee2023talkin}.
The application of these doctrines depends on many factors, most importantly on whether OpenAI has a copyright interest in its outputs. 
If these copyright doctrines do not apply, then it is still possible that publishing the dataset constitutes a breach of contract by the dataset developers.
While it would be more challenging for OpenAI to pursue a case against third parties, there are myriad other business torts, from unfair competition to misappropriation, that may be relevant to this situation, and which go beyond the scope of this paper~\citep{marks2023law}.
Time will tell the extent to which OpenAI and other LLM service providers can enforce their terms of use against third parties.
However, a prominent researcher at Google has already resigned citing concerns that OpenAI outputs were used to train BARD~\citep{VictorEfrati2023} 
In light of these legal ambiguities, our tool gives developers the ability to exclude OpenAI-generated datasets.

\vspace{-2mm}
\paragraph{While legal issues remain ambiguous, practitioners are making decisions on data use and model training.}
In the face of these pervasive legal uncertainties, practitioners' decisions regarding data usage are ultimately guided by a blend of factors including the specific licensing terms, the origin of datasets, and the degree of usage of a given dataset by others. 
Navigating this landscape requires striking a delicate balance between risk mitigation and the need for sufficient resources. 
This equation, however, varies across regions, applications, and corporate environments, influenced by factors such as competition, risk, and regional legislation.
A strategy for partially mitigating these uncertainties is for model providers to indemnify users, as done by Google Cloud~\cite{Suggs2023}. However, this may not be feasible for resource-constrained developers and, while it protects end-users, it does not solve the issues faced by model developers or dataset curators.

\vspace{-3mm}
\paragraph{Our Approach.}
The fundamental purpose of copyright is to encourage creativity and innovation.
As we highlighted in the sections above, the current legal landscape remains ambiguous and this lack of clarity can stifle innovation as developers fear legal repercussions.
Through our audit and tooling, we seek to provide important information for practitioners to make informed decisions in an otherwise ambiguous landscape, guided by their own own legal interpretation and risk tolerance.
This information includes data license lineages, a categorization of license terms, details on data creators, and the underlying data sources (e.g. web or LLM).
In light of ongoing litigation and a lack of legal certainty, we attempted to give developers 
In creating a repository of data licensing information, we are also taking a step towards encouraging dataset creators to be more thoughtful about the licenses that they select.
Dataset creators are well-positioned to understand the appropriate uses of the datasets they publish and licenses can be a tool to communicate these restrictions and to encourage responsible AI development.
We further aim to highlight that machine learning practitioners should take dataset license terms seriously, as they may have real impacts on how their models may be used in practice.
Ultimately, thoughtful data licensing could be leveraged to promote more responsible, inclusive, and transparent machine learning practices. 

\vspace{-3mm}
\textcolor{red}{\paragraph{NOTICE: Collected License Information is NOT Legal Advice.}\label{sec:not-legal-advice}
It is important to note we collect \textit{self-reported} licenses, and categorize them according to our best efforts, as a volunteer research and transparency initiative.
The information provided by any of our works and any outputs of the Data Provenance Initiative do not, and are not intended to, constitute legal advice; instead, all information, content, and materials are for general informational purposes only.
Readers and users should seek their own legal advice from counsel in their relevant jurisdiction.}

\vspace{-2mm}
\section{Related Work} 

\paragraph{Data Documentation}
A long line of work has highlighted the importance of data and its documentation in natural language processing \citep{paullada2021data, rogers-2021-changing, Meyer2023TheDM, gururangan-etal-2018-annotation,muennighoff2023scaling}.
In particular, these works stress the challenges posed by poor documentation to reproducibility, good science, and generally well-understood model behavior~\citep{Sambasivan2021EveryoneWT, bandy2021addressing, longpre2023pretrainers}.
Recent work has also explored the importance of documenting AI ecosystems~\citep{Bommasani2023EcosystemGT} and the supply chain from data to models~\citep{cen2023aisupply}.

\vspace{-2mm}
\paragraph{Data Analysis and Exploration}
Several notable works have conducted large-scale analyses into data, particularly pretraining text corpora~\citep{gao2020pile, dodge2021documenting, kreutzer2022quality, NEURIPS2022_ce9e92e3, scao2022bloom,scao2022language,mcmillan2022documenting}. 
Other works have investigated the geo-diversity of vision-based datasets~\citep{shankar2017no,de2019does,mahadev2021understanding}.
Different forms of data governance have been proposed to centralize responsibility and documentation over datasets, including for the BigScience project~\citep{jernite2022data} and a Public Data Trust~\citep{chan2023reclaiming}.
In terms of finding and visualizing datasets, a few recent tools have been proposed~\citep{farber2021datahunter, viswanathan2023datafinder}.

\vspace{-2mm}
\paragraph{Transparency and accountability}
Adjacent to the realm of legality, prior works have strongly advocated and provided frameworks for documentation and audits to increase transparency and accountability in AI systems \citep{Miceli2022DocumentingDP, Kapoor2023REFORMSRS, raji2022actionable}.
In a manner akin to DPI, which draws upon the collective knowledge of legal and machine learning experts, earlier research has also underscored the significance of interdisciplinary collaborations \citep{Hutchinson2020TowardsAF}.
Datasheets for datasets \citet{gebru2021datasheets} and Data Statements \citet{bender-friedman-2018-data} both provide structured frameworks for revealing essential metadata such as the motivation behind intended use.
\citet{pushkarna2022data} expanded on datasheets with ``Data Cards'' for sources, collection, ethics, and adoption.

Similarly, \citet{mitchell2019model} introduced model cards to benchmark model performance across demographic groups and disclose evaluation procedures.
\citet{crisan2022interactive} proposed interactive model card as an alternative mode of documentation and metadata sharing.
Complementary to transparency regarding the dataset's creation process, \citet{Corry2021ThePO} provide a framework that guides users on how to navigate datasets as they approach the end of their life-cycle.
DPI builds upon the foundational frameworks laid out in these earlier studies, with a specific focus on addressing the licensing aspects of dataset curation.
Our goal is to equip users with a comprehensive understanding of the legal risks associated with dataset usage.

\vspace{-2mm}
\paragraph{Dataset legality}
The legality of the datasets used to train large base models has recently received significant attention \citep{Sag2020TheNL,henderson2023foundation}.
The challenge of determining the legality of employing different datasets becomes particularly complex due to the intricate nature of dataset creation processes.
~\citet{lee2023talkin} break up the stages of dataset creation and model generation and assess the relevant copyright questions in the US legal system.
These processes often involve multiple licenses and restrictions that can interact in ways that obscure the final legal risk.
\citet{Soh2021BuildingLD} propose a high-level framework for pinpointing the areas within dataset creation and usage where legal analysis is necessary, but do not apply this framework to any existing datasets.
\citet{Min2023SILOLM} demonstrate that refraining from training on copyrighted or highly restricted datasets has a detrimental impact on downstream performance.
Their proposed solution involves using a language model trained on "low-risk" text and augmenting it with a data-store containing "high-risk" text which can be modified appropriately as the legal landscape clarifies over time.
~\citep{lee2023talkin}
DPI enhances these investigations by involving legal experts in the development of a framework for assessing a dataset's ``risk'' and annotating the ``risk'' associated with numerous existing high-profile datasets.

\section*{Acknowledgements}
We would like to thank Katherine Lee, A. Feder Cooper, Peter Henderson, Aviya Skowron and Stella Biderman for valuable comments and feedback.

\bibliographystyle{plainnat}
\bibliography{references}

\clearpage
\appendix
\phantomsection
\addcontentsline{toc}{section}{Appendix} 
\part{Appendix} 
\parttoc

\vspace{-3mm}
\section{Contributors}

Here we enumerate the author contributions.
We would like to emphasize that all authors contributed crucial elements to this project, and \emph{Core Contributors} in particular are recognized with hands on service to the design and construction of Data Provenance's first implementation.
\begin{itemize}
    \item \textbf{Shayne Longpre} \; Core Contributor. Primary designer and coder of the repository and explorer interface. Led audit implementation, and analysis, as well as the manual annotation process.
    \item \textbf{Robert Mahari} \; Core Contributor. Led the legal analysis, and licensing annotation design.
    \item \textbf{Anthony Chen} \; Core Contributor. Led automatic inferencing of dataset text metrics, topics, and task category annotations. Supported writing, analysis, and code testing.
    \item \textbf{Naana Obeng-Marnu} \; Core contributor. Led visualization design, particularly interactive visualizations in the Data Provenance Explorer.
    \item \textbf{Damien Sileo} \; Core contributor. Led data aggregator linking, and metadata scraping. Supported writing, analysis, source annotation and adding datasets.
    \item \textbf{William Brannon} \; Core contributor. Added 8 data collections, supported writing and data analysis.
    \item \textbf{Niklas Muennighoff} \; Core contributor. Added several large data collections, supported writing, analysis, visualization, and source annotations.
    \item \textbf{Nathan Khazam} \; Core contributor. Led licensing annotation effort and supported adding datasets along with testing.
    \item \textbf{Jad Kabbara} \; Core contributor and advisor. Led text source annotation effort and supported with framing, writing and analysis.
    \item \textbf{Kartik Perisetla} \; Core contributor. Added several datasets, supported writing, analysis, and dataset preparation for Hugging Face.
    \item \textbf{Xinyi (Alexis) Wu} \; Core contributor. Added several datasets, testing, and supported automatic metadata collection.
    \item \textbf{Enrico Shippole} \; Core contributor. Led final dataset preparation for Hugging Face upload and testing.
    \item \textbf{Kurt Bollacker} \; Advisor on project design and framing. 
    \item \textbf{Tongshuang Wu} \; Advisor, particularly on data analysis and visualizations. Supported writing and Data Provenance Explorer design.
    \item \textbf{Luis Villa} \; Advisor on data copyright and licensing, and supporting writing in the legal discussion section.
    \item \textbf{Sandy Pentland} \; Advisor on general project design and framing.
    \item \textbf{Sara Hooker} \; Advisor on general project design and framing, as well as supporting writing, analysis, and directing experiments.

\end{itemize}


\section{Exact Licenses and Citations}

See \cref{tab:licenses} for a summary of the Data Provenance Collection licenses and citations.
More comprehensive details are available at \url{https://github.com/Data-Provenance-Initiative/Data-Provenance-Collection}.

\begin{table*}
\centering
\begin{tabular}{l|lp{5.5cm}}
\toprule
 & Cite & Licenses \\
Collection &  &  \\
\midrule
Airoboros & \citet{Durbin2023Airoboros} & CC BY-NC 4.0 \\
Alpaca & \citet{alpaca} & CC BY-NC 4.0 \\
Anthropic HH & \citet{bai2022training, gangulired} & MIT License \\
BaizeChat & \citet{xu2023baize} & CC BY-NC 4.0 \\
BookSum & \citet{kryscinski2022booksum} & Academic Only \\
CamelAI Sci. & \citet{li2023camel} & CC BY-NC 4.0 \\
CoT Coll. & \citet{kim2023cot} & Non Commercial \\
Code Alpaca & -- & Unspecified \\
CommitPackFT & \citet{muennighoff2023octopack} & Various \\
Dolly 15k & \citet{dolly15k_2023} & CC BY-SA 3.0 \\
Evol-Instr. & \citet{xu2023wizardlm} & Academic Only \\
Flan Collection & \citet{longpre2023flan} & Various \\
GPT-4-Alpaca & \citet{peng2023instruction} & CC BY-NC 4.0 \\
GPT4AllJ & \citet{gpt4all} & Various \\
GPTeacher & -- & Unspecified \\
Gorilla & \citet{patil2023gorilla} & Apache License 2.0 \\
HC3 & \citet{guo2023close} & Various \\
Joke Expl. & -- & MIT License \\
LAION OIG & \citet{oig2023} & Various \\
LIMA & \citet{zhou2023lima} & CC BY-NC-SA 4.0 \\
Longform & \citet{koksal2023longform} & CC BY-SA 3.0, Unspecified, CC BY-SA 4.0 \\
OpAsst OctoPack & \citet{muennighoff2023octopack} & CC BY 4.0 \\
OpenAI Summ. & \citet{stienon2020learning} & CC BY 4.0 \\
OpenAssistant & \citet{kopf2023openassistant} & CC BY 4.0 \\
OpenOrca & \citet{mukherjee2023orca} & Various \\
SHP & \citet{SHP} & Unspecified \\
Self-Instruct & \citet{selfinstruct2022} & Apache License 2.0 \\
ShareGPT & \citet{sharegpt} & Unspecified \\
StackExchange & -- & Unspecified \\
StarCoder & \citet{li2023starcoder} & BigScience OpenRAIL-M \\
Tasksource Ins. & \citet{sileo2023tasksource} & Various \\
Tasksource ST & \citet{weston2015aicomplete} & Various \\
TinyStories & \citet{eldan2023tinystories} & CDLA Sharing 1.0 \\
Tool-Llama & \citet{qin2023toolllm} & CC BY-NC 4.0 \\
UltraChat & \citet{ding2023enhancing} & CC BY-NC 4.0 \\
Unnatural Instr. & \citet{honovich2022unnatural} & MIT License \\
WebGPT & \citet{nakano2021webgpt} & Apache License 2.0, CC BY-SA 4.0 \\
xP3x & \citet{muennighoff2022crosslingual} & Various \\
\bottomrule
\end{tabular}
\caption{\textbf{Licenses and citations} for the dataset collections presented in this paper. Collections containing material under more than three distinct licenses are marked as having ''Various`` licenses, and we refer readers to our raw data for the full details.}
\label{tab:licenses}
\end{table*}

\vspace{-1mm}
\section{Details on Collecting Data Provenance}
\label{sec:data-prov-collection}

This data was collected with a mix of manual and automated techniques, leveraging dataset aggregators like GitHub, Hugging Face and Semantic Scholar.
Annotating and verifying license information, in particular, required a carefully guided manual workflow, designed with legal practitioners (see~\cref{sec:license-collection}).
Once these information aggregators were connected, it was possible to synthesize or scrape additional metadata, such as dataset languages, task categories, and time of collection.
And for richer details on each dataset, like text topics and source, we used carefully tuned prompts on language models inspecting each dataset.

\vspace{-1mm}
\paragraph{Automated Annotation Methods}
Based on the manually retrieved pages, we automatically extract Licenses from HuggingFace configurations and GitHub pages. We leverage the Semantic Scholar public API \citep{Kinney2023TheSS} to retrieve the released date and current citation counts associated to academic publications.
Additionally, we compute a series of other helpful, but often overlooked data properties such as text metrics (the min/mean/max for input and target lengths), and dialog turns.
We elected to measure sequence length in characters rather than word tokens, for fairer treatment across language and script given well-known differences in tokenizer performance across different languages \citep{petrov2023language}.

\vspace{-1mm}
\paragraph{API Annotation Methods}

While Task Categories have become the established measurement of data diversity in recent instruction tuning work \citep{sanh2021multitask, selfinstruct2022}, there are so many other rich features describing data diversity and representation.
To augment this, we use OpenAI's GPT-4 API to help annotate for text topics.
We randomly sampled 100 examples per dataset and carefully prompt GPT-4 to suggest up to 10 topics discussed in the text.

To annotate for the original data sources, AI experts (PhD students and postdocs) reviewed the papers and filled out the original text sources, whether machines or template-generation were used for synthetic generation, and whether human annotators were used.
GPT-4 was used as an in-context retriever on the dataset's ArXiv paper to extract snippets that the experts may have missed. 
We split the ArXiv paper into 4000 characters chunks and prompt the API to return a json list of any mentions of the dataset source, e.g. from scraping, synthetic or manual generation.

\end{document}